\documentclass[runningheads]{llncs}

\usepackage{eccv}

\usepackage{eccvabbrv}

\usepackage{graphicx}
\usepackage{booktabs}

\usepackage[accsupp]{axessibility}  

\usepackage[pagebackref,breaklinks,colorlinks,citecolor=eccvblue]{hyperref}

\usepackage{orcidlink}

\newcommand{\modelname}{InstructAvatar\xspace}
\newcommand{\update}{\textcolor{black}}
\usepackage{multirow}
\usepackage{tcolorbox}

\begin{document}

\title{InstructAvatar: Text-Guided Emotion and Motion Control for Avatar Generation} 

\titlerunning{InstructAvatar}

\author{Yuchi Wang \and
Junliang Guo \and
Jianhong Bai\and
Runyi Yu\and
Tianyu He\and
Xu Tan\and \\
Xu Sun\and
Jiang Bian}
\authorrunning{Yuchi Wang et al.}

\institute{Peking University \\
\email{wangyuchi@stu.pku.edu.cn}\\
~\\
\url{https://wangyuchi369.github.io/InstructAvatar/}}

\maketitle

\begin{figure}
    \centering
    \includegraphics[width=0.9\textwidth]{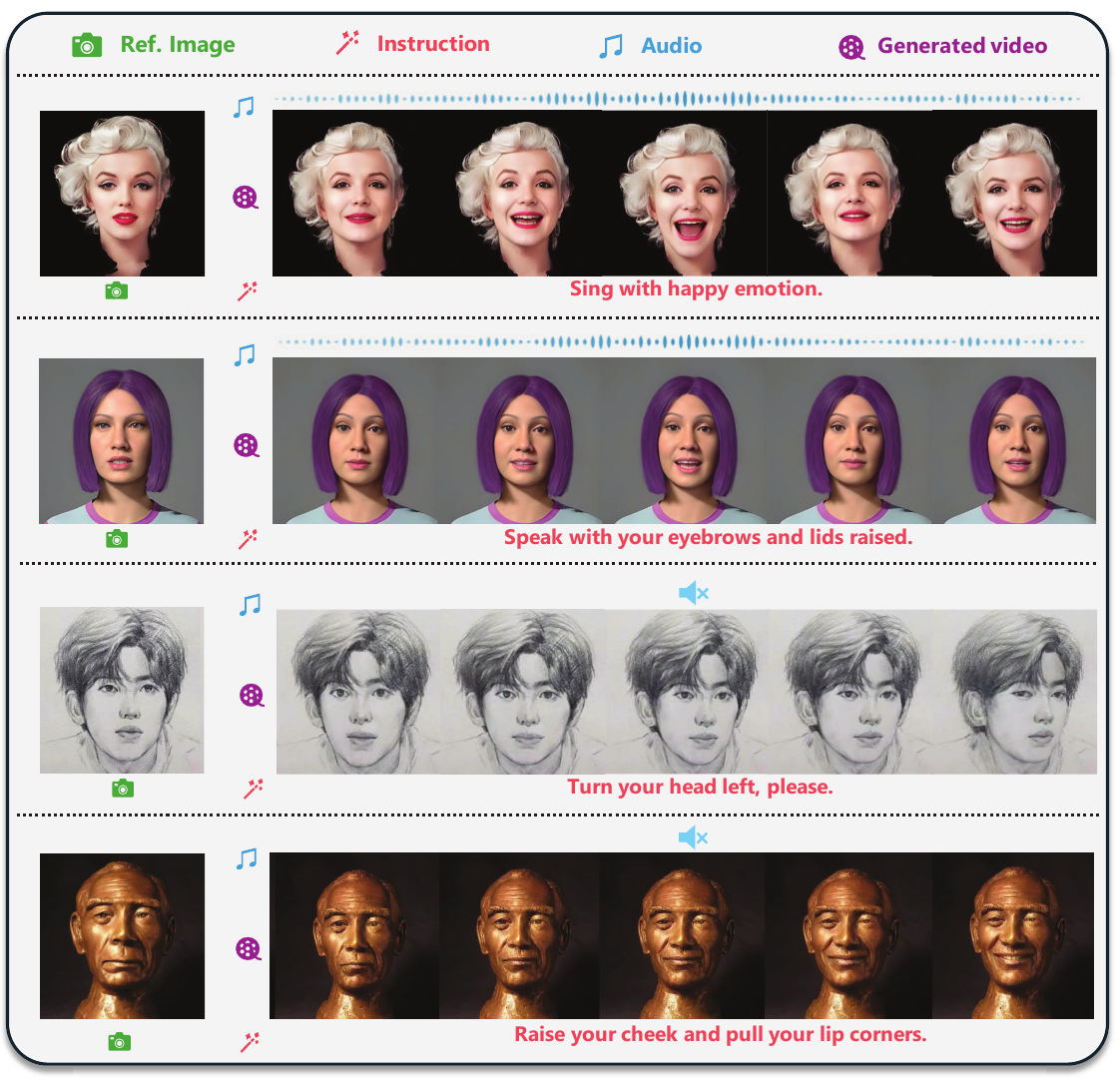}
    \caption{
        \modelname enables emotional talking face generation through a flexible natural language interface (top 2 rows). The generated results exhibit fine-grained expression control, excellent identity preservation, high-quality lip sync, and natural character movements. Moreover, it supports direct control of facial motion and expression without relying on audio cues, a feature absent in previous studies (bottom 2 rows). The ability of \modelname to handle highly out-of-domain appearances (like cartoons, sketches, and sculptures) further highlights its generalization capabilities. }
    \label{fig:demo}
\end{figure}

\begin{abstract}

Recent talking avatar generation models have made strides in achieving realistic and accurate lip synchronization with the audio, but often fall short in controlling and conveying detailed expressions and emotions of the avatar, making the generated video less vivid and controllable. 
In this paper, we propose a novel text-guided approach for generating emotionally expressive 2D avatars, offering fine-grained control, improved interactivity and generalizability to the resulting video. Our framework, named \modelname, leverages a natural language interface to control the emotion as well as the facial motion of avatars. Technically, we design an automatic annotation pipeline to construct an instruction-video paired training dataset, equipped with a novel two-branch diffusion-based generator to predict avatars with audio and text instructions at the same time. 
Experimental results demonstrate that \modelname produces results that align well with both conditions, and
outperforms existing methods in fine-grained emotion control, lip-sync quality, and naturalness.

\end{abstract}

\section{Introduction}
\label{sec:intro}

Avatar generation has recently gained significant attention due to its broad applicability in film production, gaming, video conferencing, and various other domains. The primary objective of this technology is to animate portraits with synchronized speech audio. While previous studies have achieved impressive lip synchronization and head pose prediction~\cite{he2023gaia, tian2024emo, sun2023vividtalk, zhang2023sadtalker, wang2021audio2head, zhou2020makelttalk, chu2024gpavatar, wang2022one}, effectively conveying and controlling detailed expressions and motions remains a challenge, resulting in less vivid and authentic videos.
Previous studies have attempted to integrate emotional information through either labels~\cite{wang2020mead, eat, gururani2023space, zhai2023talking, emmn} or example videos~\cite{ma2023dreamtalk, ma2023styletalk, pd-fgc, ji2022eamm, zhang2023dream}. However, they continue to face challenges related to limited flexibility and controllability.

As a seamless interface bridging the gap between humans and computers, the textual prompt stands out as a natural solution to tackle the limitations of previous approaches, providing a versatile array of controls that encompass desired motions and expressions for the avatar.
Therefore, we steer towards a textual instruction-based talking avatar generation model named \modelname, which offers three key advantages:  (1) Enhanced control over fine-grained details rather than just the overall style; (2) Improved generalizability compared to limited emotion or style categories; and (3) Enhanced interactivity and user-friendliness. 
As illustrated in~\cref{fig:demo}, our framework facilitates text-guided emotional talking face generation with fine-grained control, while enabling facial motion/expression control without relying on audio cues.

To accomplish this, we meticulously design our algorithm, taking into account both data and model architecture considerations. For data preparation, existing datasets~\cite{wang2020mead, cao2014crema} only offer tag-level emotion annotations.
 To capture fine-grained facial details, we utilize Action Units (AUs)\cite{ekman1978facial} to describe facial muscle movements. AUs are extracted from each video clip using an off-the-shelf model\cite{audetect}. Subsequently, we prompt a large-scale multi-modal language model GPT-4V~\cite{gpt4v} to paraphrase AUs into a natural textual description. Additionally, leveraging the visual capabilities of GPT-4V, we refine the detected AUs and provide additional facial information. As a result, we generate a dataset containing detailed emotion and motion descriptions.
Regarding the model architecture of \modelname, 
we propose a diffusion model with two branches of cross-attentions to incorporate different types of instructions while generating the talking video, i.e., the emotion instructions that are high-level throughout the entire video, and the facial motions that are dynamic over timestamps.
Additionally, novel techniques such as zero-convolution gate are proposed to stabilize the training and enhance the guidance,
which will be detailed in~\cref{sec: method}.

For experiments, we propose several tailored evaluation metrics to justify the model's performance on fine-grained facial emotion and motion control. Experimental results demonstrate that: (1) \modelname exhibits significant improvements in emotion control, lip-sync quality, and naturalness compared to previous baselines. (2) Notably, our model has a natural language interface, enabling it to receive a much wider range of instructions. (3) Additional experimental results indicate that our model can, for the first time, effectively animate avatars directly without audio. In summary, the contributions of our paper are as follows:

\textbf{(1)} We introduce \modelname, a diffusion-based avatar generation model featuring a fine-grained natural language instruction interface. It showcases superior flexibility, control effectiveness, and naturalness compared to previous methods. To our best knowledge, it is the first text-guided 2D-based talking face generation framework.

\textbf{(2)} Through meticulous design, like the implementation of a two-branch cross-attention mechanism, we integrate text-guided facial motion control into our unified framework, further enhancing the scope of avatar control.

\textbf{(3)} We annotate an instruction-video dataset and establish an evaluation pipeline for the fine-grained emotional talking video generation task, which may facilitate further research.

\section{Related Works}

The rapid advancement of generative models such as GANs~\cite{goodfellow2014generative} and diffusion models~\cite{ho2020denoising} has led to remarkable progress in avatar generation. Several studies have explored the direct generation of facial avatars \cite{zhang2023dreamface, li2023instruct, liu2024towards, canfes2023text, yu2023towards}, as well as the editing of human face images \cite{xia2021tedigan, patashnik2021styleclip, wang2024facecomposer, huang2023collaborative}. However, these works often produce static images without dynamic effects. In contrast, talking head generation, a fundamental task in avatar generation, aims to generate a video in which an avatar speaks the provided audio. This task can be broadly categorized into video-driven and audio-driven approaches. In video-driven methods~\cite{siarohin2019first, wang2021facevid2vid, zhang2022metaportrait, tripathy2021facegan}, the movement of a portrait is generated based on another driving video, while in audio-driven methods~\cite{zhang2023sadtalker, zhou2020makelttalk, tian2024emo, he2023gaia}, motion is predicted directly from audio inputs. To tackle the challenge of learning facial motion representations, previous talking head models have often relied on domain priors like warping-based transformations \cite{wang2021audio2head, gururani2023space, zhou2020makelttalk, liu2022audio} or 3D Morphable Models \cite{zhang2021flow, ren2021pirenderer, zhang2023sadtalker}. Recently, \cite{he2023gaia} proposed a disentangled motion and appearance architecture and collected a large-scale dataset to directly learn the data distribution, thereby further enhancing the naturalness and diversity of the generated avatars.


Acknowledging the constraints of prior efforts that often yield emotionless avatars, there has been growing interest in injecting emotions into talking face generation. For instance,  MEAD~\cite{wang2020mead} represents emotion using a one-hot vector, while EAT \cite{eat} employs a mapping network to extract emotion guidance through a latent code. EAMM~\cite{ji2022eamm} represents the facial dynamics of reference emotional video as displacements to motion representations. PD-FGC~\cite{pd-fgc} disentangles control over specific facial organs with emotional expression, and StyleTalk \cite{ma2023styletalk} develops a style encoder to extract the style of a reference video. 
These methods either support a limited range of coarse emotion types~\cite{wang2020mead, eat, gururani2023space, zhai2023talking, emmn} or necessitate users to seek out another desired style video~\cite{ma2023dreamtalk, ma2023styletalk, pd-fgc, ji2022eamm, zhang2023dream}, limiting the flexibility and controllability of the generated avatars.

More recently, some endeavors have aimed to incorporate text as an emotion control signal~\cite{zhao2024media2face, xu2023high, zhong2023expclip, ma2023talkclip}. However, they typically utilize text to generate emotional talking 3D animations, which requires an external renderer to convert these animations into real talking videos. Although this animation-based approach may alleviate difficulty by introducing more domain priors, it inherently leads to indirect controls
and restricted diversity \cite{he2023gaia}.

In this paper, we propose to directly learn the distribution of talking videos and enable fine-grained controlling with textual prompts, improving the naturalness and controllability of generation results.



\section{Methodology} \label{sec: method}

In this section, we introduce the \modelname model. In~\cref{sec: model_overview}, we provide an overview of the architecture of \modelname. Subsequently,~\cref{sec: method_data} describes the process of constructing high-quality text instructions, while~\cref{sec: method_arch} delineates each component of our \modelname. Lastly, in~\cref{sec: method_pipeline}, we demonstrate the training and inference pipelines.

\subsection{Overview} \label{sec: model_overview}
\begin{figure}
    \centering
    \includegraphics[width=1\textwidth]{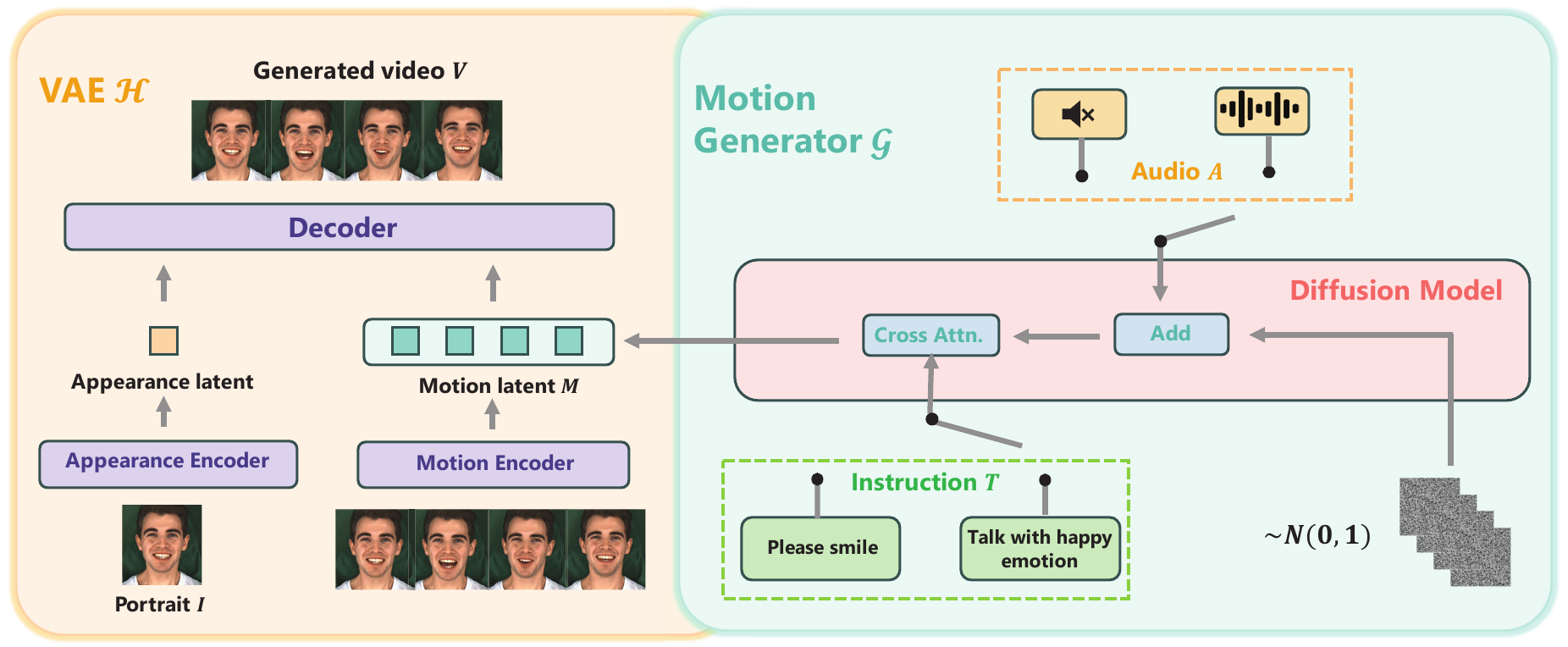}
    \caption{
Method Overview: The \modelname consists of two components: VAE $\mathcal{H}$ to disentangle motion information from the video and a motion generator $\mathcal{G}$ to generate the motion latent conditioned on audio and instruction. As we have two types of data, two switches in instruction and audio are designed. During inference, the motion encoder in the VAE will be dropped and we iteratively denoise Gaussian noise to obtain the predicted motion latent. Together with the user-provided portrait, the resulting video is generated by the decoder of the VAE.}
    \label{fig:overview}
\end{figure}

Given a sequence of audio clips $\boldsymbol{A}=[a_1, a_2, ..., a_N]$, one portrait image $\boldsymbol{I}$, and the text instruction $\boldsymbol{T}$, our model is tasked with animating the portrait to utter the audio with the target style represented by the instruction. In other words, we aim to learn a mapping to generate a video $\boldsymbol{V}=\mathcal{F}(\boldsymbol{A}, \boldsymbol{I}, \boldsymbol{T})$.

As illustrated in~\cref{fig:overview}, we decompose $\mathcal{F}$ into two parts: variational autoencoder (VAE)  $\mathcal{H}$ and diffusion-based motion generator $\mathcal{G}$. The VAE follows the approach outlined in\cite{he2023gaia} to disentangle motion information from appearance, which means that we can derive $\boldsymbol{V}=\mathcal{H}(\boldsymbol{M}, \boldsymbol{I})$, where $\boldsymbol{M}$ represents purely motion information and $\boldsymbol{I}$ is the provided portrait. More details about VAE and motion latent $\boldsymbol{M}$ can be found in the~\cref{sec:app_vae}. Now, we can focus on learning the motion generator conditioned on audio and textual instructions, \ie, $\boldsymbol{M}=\mathcal{G}(\boldsymbol{A}, \boldsymbol{T})$. 

In the following sections, we will detail how to obtain fluent, diverse, and fine-grained text instructions $\boldsymbol{T}$, as well as how to design a diffusion model-based text-guided motion generator $\mathcal{G}$.



\subsection{Construct Natural and Diverse Text Instructions} \label{sec: method_data}

\begin{figure}
    \centering
    \includegraphics[width=1.05\textwidth]{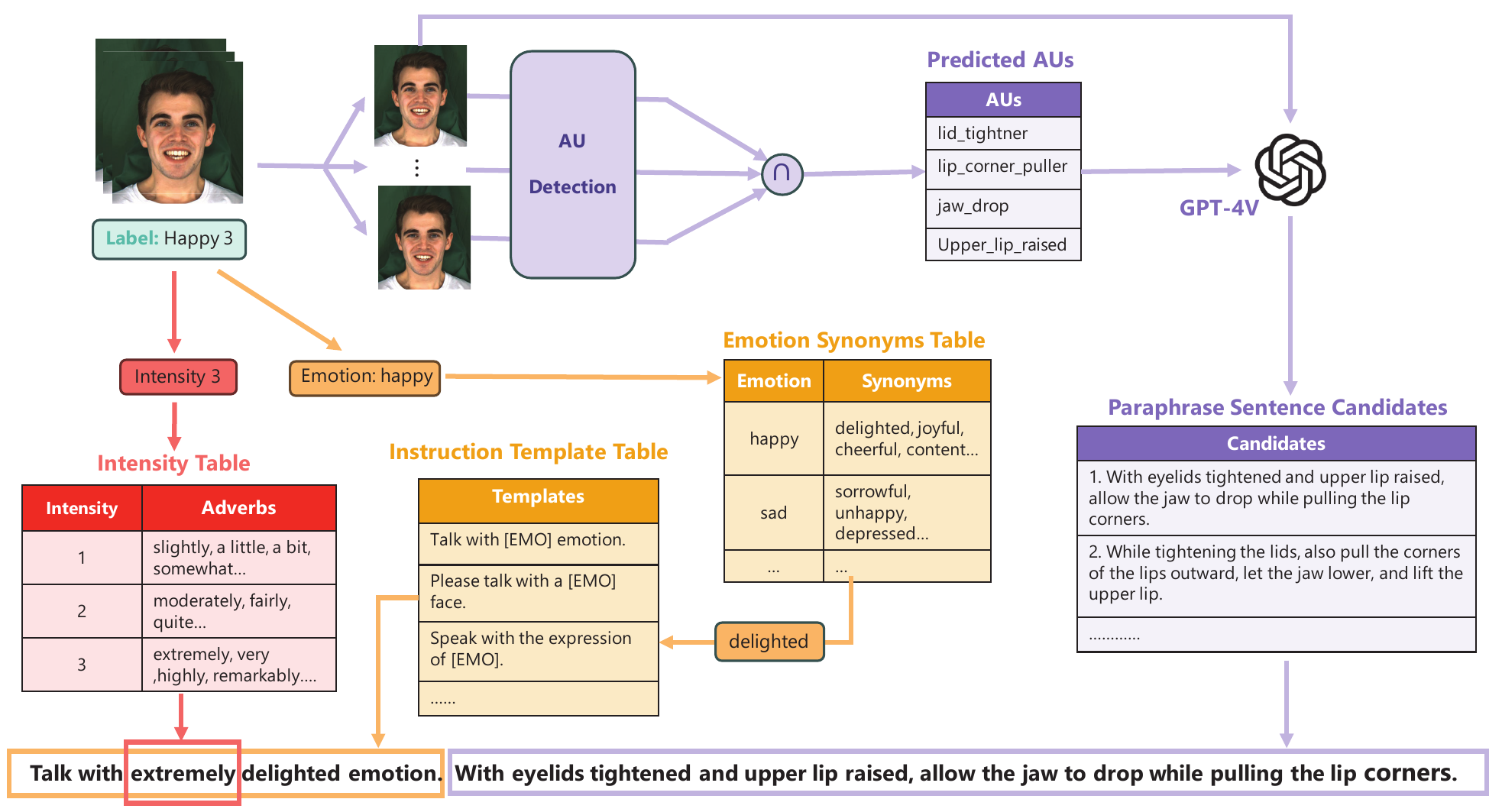}
    \caption{We extend the emotion label using a predefined template and incorporate intensity information by modifying the emotion with an adverb representing the degree. For fine-grained control, we extract the AUs and then prompt GPT-4V to paraphrase them into a sentence. }
    \label{fig:data}
\end{figure}

To utilize natural language as the interactive interface, it is essential to construct a dataset containing text-expression pairs. However, existing emotional talking datasets~\cite{wang2020mead, cao2014crema} typically provide only tag-level annotations for talking videos, offering limited emotion categories such as ``happy'' or ``sad'' along with their corresponding videos. To diversify and refine these labels into text instructions, we will employ three procedures: Emotion Label Extension, Action Unit Extraction, and MLLM Paraphrase, aiming to generate diverse, fine-grained and natural text instructions $\boldsymbol{T}$, illustrated in~\cref{fig:data}.

\subsubsection{Emotion Label Extension}

Firstly, we adopt a straightforward and convenient method to convert emotion labels into sentences by utilizing predefined templates.
Specifically, we prompt ChatGPT to generate 60 templates, such as ``Talk with \texttt{[EMO]} emotion''. 
We then randomly select a synonym from a predefined table and then substitute the placeholder \texttt{[EMO]} in the template, producing expressions like ``Talk with delighted emotion''.
Additionally, we also utilize the emotion intensity information of the dataset in our training. For instance, for videos with high emotion intensity, we may add adverbs like ``extremely'' to modify emotions, resulting in ``extremely delighted''. 

\subsubsection{Action Unit Extraction}

The previous method for emotion label-based extension tends to provide coarse and high-level annotations. To enable fine-grained control of facial expressions, we turn to the Facial Action Coding System (FACS)\cite{ekman1978facial}. This system defines Action Units (AUs) to describe facial muscle movements, allowing for a detailed description of the local states of a talking face as shown in~\cref{fig:data}. However, AU detection is typically performed on images. Therefore, we randomly select three frames from a video and employ an off-the-shelf AU detection model\cite{audetect} to extract AUs from these images. We then take the intersection of predicted action units, considering that significant facial states corresponding to a specific emotion are likely to be consistent throughout the entire video.

\subsubsection{MLLM Paraphrase} \label{sec: paraphrase}

The action units obtained in the previous step are represented in a relatively incomprehensible form, such as ``\texttt{lid\_tightner}''. To transform them into more human-friendly text instructions, we capitalize on large language models' powerful paraphrase capability. We prompt GPT-4V \cite{gpt4v} to combine these action units into sentences. Additionally, leveraging its vision capabilities, we provide GPT-4V with a frame extracted from the video and allow it to edit the action units if it disagrees with the action units extracted by the off-the-shelf model. We ask GPT-4V to provide a few diverse sentences and randomly select one during training. More analysis on the reliability of GPT-4V can be found in~\cref{sec:app_gpt}.

\subsection{Text-Guided Motion Generator} \label{sec: method_arch}

\begin{figure}
    \centering
    \includegraphics[width=1\textwidth]{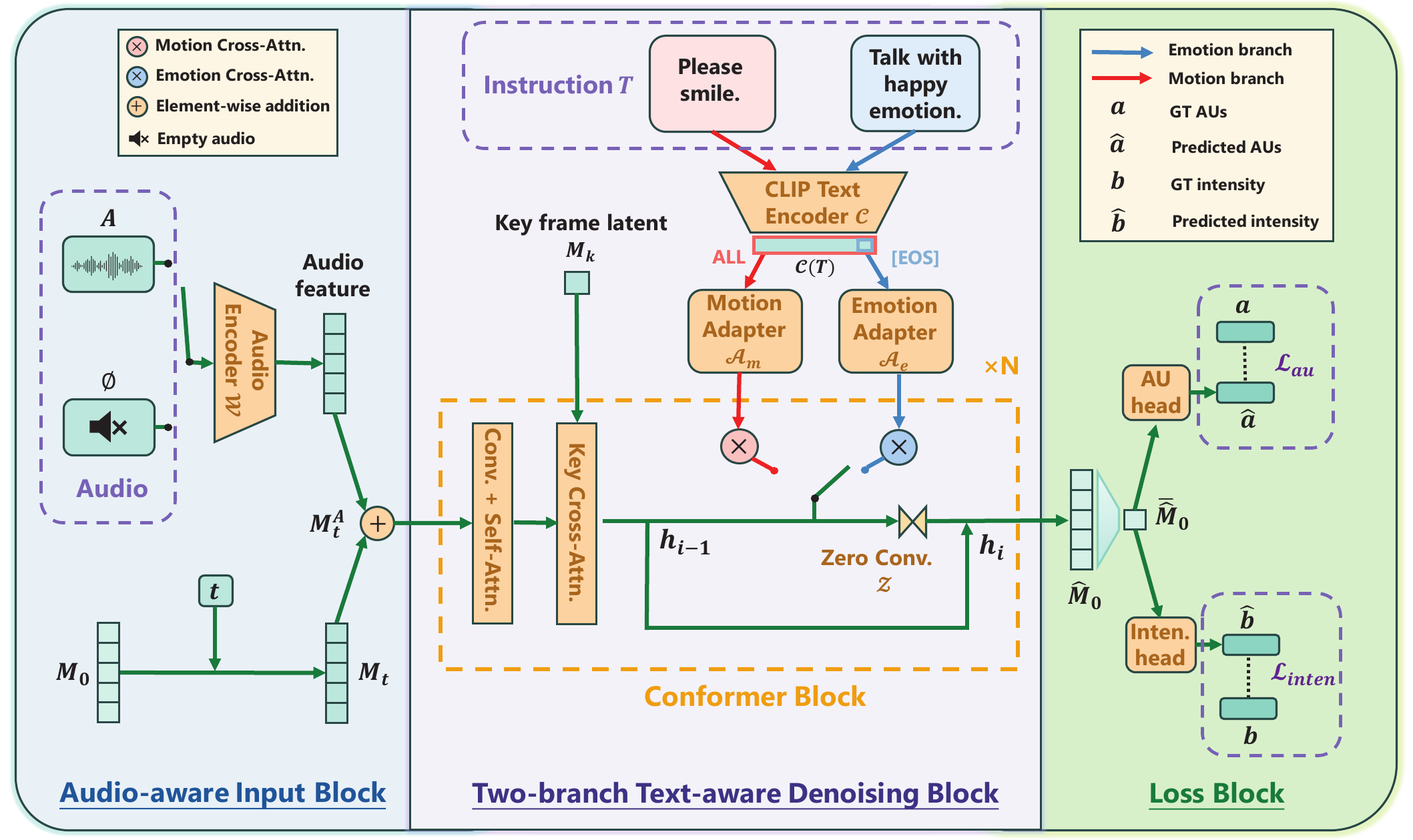}
    \caption{Details of motion generator.  To denoise noisy motion latent, firstly we element-wise add the audio feature into it. Then, in each denoising block, we design a two-branch cross-attention module to inject emotion and motion control into the model. Lastly, we also incorporate AU and intensity losses to encourage the model to learn them. }
    \label{fig:main}
\end{figure}

We leverage the diffusion model~\cite{ho2020denoising}, a powerful generative model, as our text-guided motion generator to learn $\boldsymbol{M}=\mathcal{G}(\boldsymbol{A}, \boldsymbol{T})$ mentioned in~\cref{sec: model_overview}. In the \modelname, we use Conformer~\cite{gulati2020conformer} as our diffusion model backbone. The details of our motion generator is illustrated in~\cref{fig:main}.

\subsubsection{Basics for Diffusion Models} \label{sec: method_diff_basic}
The diffusion model is designed to fit a distribution. Basically, it is divided into two phases: the forward diffusion process and the backward denoising process. Given a data point sampled from a real data distribution $x_0\sim q(x)$, we define a forward process in which Gaussian noise is incrementally added to the sample, generating a sequence of noisy samples $x_1, ..., x_T$. The noise scales are controlled by $\beta_t \in (0,1)$, and the density is expressed as
$q(x_t|x_{t-1})= \mathcal{N}(x_t; \sqrt{1-\beta_t}x_{t-1}, \beta_t\mathbf I)$.
Based on the reparameterization trick~\cite{ho2020denoising}, we can sample at any arbitrary time step in a closed form: 
$  q(x_t|x_0) = \mathcal{N}(x_t; \sqrt{\bar\alpha_t}x_0, \sqrt{1-\bar\alpha_t}\mathbf I)$, where $\alpha_t=1-\beta_t$ and $\bar{\alpha}_t=\prod_{i=1}^t \alpha_i$. 
Furthermore, from this equation, it becomes evident that as $T\to \infty$, $x_T$ converges to an isotropic Gaussian distribution.

For the reverse process, in DDPM~\cite{ho2020denoising}, it derives a simple learning objective 
$
    \mathcal \mathcal{L}_{\rm simple}=\sum_{t=1}^T\mathbb{E}_q \big[||\epsilon_t(x_t, x_0)-\epsilon_{\theta}(x_t, t)||^2 \big]
$, where $\epsilon_t$ is the noise added in original data $x_0$ and $\epsilon_\theta$ is learnable network. 
Recently, researchers tend to use an even simpler strategy to train a network to predict $x_0$ directly, with the loss function defined as $L=||x_0-f_\theta(x_t,t)||$, which is also applied in our framework. During inference, following DDIM~\cite{song2020denoising}, we start from a Gaussian noise and iteratively denoise it to get a predicted $\hat{x}_0$. 

\subsubsection{Audio-aware Input Block}


Basically, following the classical strategy of diffusion models mentioned above, we will train the denoising block to recover noised motion latent $\boldsymbol{M}_t$ to predicted $\hat{\boldsymbol{M}}_0$, denoted as $\hat{\boldsymbol{M}}_0 = f(\boldsymbol{M}_t, t, \boldsymbol{A}, \boldsymbol{T})$.

To incorporate audio information into the denoising process, we first normalize the speech to an appropriate amplitude range and then apply a denoiser \cite{defossez2020real} to reduce background noise. Subsequently, we utilize Wave2Vec 2.0  \cite{baevski2020wav2vec} as audio encoder $\mathcal{W}$ to extract audio features. As a special case, for facial motion control absent of audio, we use pseudo empty audio with zero amplitude and a length aligned with the ground truth video. Given that audio and motion sequences are aligned in the temporal semantic, we opt to element-wise add the audio features $\boldsymbol{A}$ to the motion latent vector $\boldsymbol{M}$. In summary, we obtain the audio-aware noisy latent $\boldsymbol{M}_t^{\boldsymbol{A}}$ by:
\begin{equation*}
   \boldsymbol{M}_t^{\boldsymbol{A}} = \left\{
   \begin{aligned}
    &\boldsymbol{M}_t \oplus \mathcal{W}(\boldsymbol{A}) \quad  &&\text{If performing emotional talking control.} \\
    &\boldsymbol{M}_t \oplus \mathcal{W}(\varnothing)  &&\text{If performing facial motion control.}
    \end{aligned}
    \right.
\end{equation*}


\subsubsection{Two-branch Text-aware Denoising Block}

Now we would use cascaded denoising blocks to denoise $\boldsymbol{M}_t^{\boldsymbol{A}}$ to the data $\hat{\boldsymbol{M}}_0$. A key component of our architecture is injecting text instruction information into the denoising procedure. To encode the text instructions $\boldsymbol{T}$ obtained in~\cref{sec: method_data}, we leverage the CLIP\cite{CLIP} text encoder $\mathcal{C}$, which has been proven for its powerful cross-modality alignment ability and strong semantic generalization ability. We employ a cross-attention mechanism to incorporate text information, where the hidden states in the Conformer layer act as queries, and the text representation serves as keys and values.

There exist some differences between emotion and motion controls. 
For the emotion, the text provides style guidance throughout the entire video, ensuring that the avatar maintains the desired emotion consistently. 
However, this is not the case for facial motion instructions, which usually describe gradually achieved actions and transitions over time. For example, a person may gradually turn his head when receiving corresponding motion instructions.
In response, we split the data flow into two branches in each denoising block when incorporating text information. For the emotion branch, we use the \texttt{[EOS]} token from the CLIP text encoder, which encapsulates overall information about the instruction. For the motion branch, we utilize the hidden states of all tokens from the last layer of the CLIP text encoder to capture more detailed and dynamic information. Additionally, we introduce different Adapters~\cite{gao2024clip} $\mathcal{A}_e, \mathcal{A}_m$ to better align the distributions of these two spaces with the space that facilitates the diffusion model's learning process. Overall, the instruction representation $\operatorname{Rep}(\boldsymbol{T})$ to inject into the denoising procedure for text-guided emotional talking and text-guided facial motion control could be summarized as:
\begin{equation*}
   \operatorname{Rep}(\boldsymbol{T}) = \left\{
   \begin{aligned}
    &\mathcal{A}_e (\mathcal{C}(\boldsymbol{T})_{[EOS]})  \quad &&\text{If $\boldsymbol{T}$ is emotional talking instruction.} \\
    &\mathcal{A}_m (\mathcal{C}(\boldsymbol{T})_{all})  &&\text{If $\boldsymbol{T}$ is motion control instruction.}
    \end{aligned}
    \right.
\end{equation*}

In addition, to provide the generator with facial shape information, following~\cite{he2023gaia}, we randomly select a frame from the video's motion latent $\boldsymbol{M}_0$ as the key frame latent $\boldsymbol{M}_k$ and inject it into the denoising process through cross-attention. It's noteworthy that for emotional talking, this selected key frame may inadvertently leak emotional information. Therefore, we substitute this frame with a frame from another emotional type video featuring the same person.

\subsubsection{Zero Convolution Mechanism for Text Condition} \label{sec: method_zero}
To leverage the abundant knowledge obtained from previous talking head models, we initialize our model from pretrained emotion-unaware models. Therefore, directly inserting text instructions would significantly reduce the expressiveness of the previously learned parameters. An insight is that an emotional talking video could transform from a neutral expression video gradually. Therefore, inspired by \cite{zhang2023adding}, we tailor a zero convolution mechanism for text conditioning.

Specifically, suppose we have hidden states $h_{i-1}$ before entering the cross-attention module with text instructions. We then use \[h_i = h_{i-1} + \mathcal{Z}(\text{Cross-Attn}(h, \text{Rep}(\boldsymbol{T})))\]
to get the next hidden states. In this formula, $\mathcal{Z}$ is a zero convolution operation where a 1-dimensional convolutional kernel moves along the hidden states dimension, with both the weight and bias initialized to zero. Therefore, at the start of training, $h_i = h_{i-1}$, which corresponds to the no instruction setting. Consequently, the zero convolution layer serves as a gate to slowly inject text control instructions into the pre-trained talking face model, stabilizing the training process and leveraging the abundant knowledge obtained in previous emotion-unaware models.

\subsection{Training and Inference Pipelines} \label{sec: method_pipeline}

\subsubsection{Loss Definition}
After cascaded denoising blocks, we obtain the predicted motion latent $\hat{\boldsymbol{M}}_0$. The most intuitive loss is the distance between the predicted motion latent and the ground truth $\boldsymbol{M}_0$, expressed as $\mathcal{L}_{mse} = \|\hat{\boldsymbol{M}}_0 - \boldsymbol{M}_0\|_2^2$. Additionally, to enforce the model to pay attention to action units and emotion intensity, we jointly train two classifier heads. We perform mean pooling in the temporal dimension followed by two-layer MLPs to extract information from $\hat{\boldsymbol{M}}_0$. $\mathcal{L}_{au}$ is calculated using the binary cross-entropy (BCE) loss, treating it as a multi-label classification problem over predicted AU logits $\hat{\boldsymbol{a}} \in \mathbb{R}^M$ and ground-truth labels $\boldsymbol{a}$, given by: $\mathcal{L}_{au} = -\frac{1}{N}\sum_{i=1}^N \|\boldsymbol{a} \odot \log(\hat{\boldsymbol{a}}) + (1-\boldsymbol{a}) \odot \log(1-\hat{\boldsymbol{a}})\|_1$, where $M$ is the number of action units and $N$ is the sample number.  For the emotion intensity loss, which is a three-classification problem, we use the standard cross-entropy loss $\mathcal{L}_{inten} = -\frac{1}{N}\sum_{i=1}^N \boldsymbol{b} \log(\hat{\boldsymbol{b}})$, where $\hat{\boldsymbol{b}} \in \mathbb{R}^3$ represents predicted logits and $\boldsymbol{b}$ represents the corresponding label. Moreover, following the approach in \cite{he2023gaia}, we additionally train a head pose predictor and use another mean squared error (MSE) loss $\mathcal{L}_{pose}=\|\hat{\boldsymbol{P}} - \boldsymbol{P}\|_2^2$ to measure the predicted pose and the ground truth. In summary, our loss is defined as :
\[L = \mathcal{L}_{mse} + \lambda_{pose} \mathcal{L}_{pose} +\lambda_{au} \mathcal{L}_{au} + \lambda_{inten} \mathcal{L}_{inten} \]
where $\lambda_{pose}$, $\lambda_{au}$, $\lambda_{inten}$ are hyperparameters.

\subsubsection{Inference Pipeline}

During inference, we begin by sampling a Gaussian noise $\boldsymbol{M}_T$ to initialize the motion latent. The audio and instructions are provided by the user. We employ the VAE motion encoder~\cite{he2023gaia} to encode the user-provided portrait $\boldsymbol{I}$, resulting in the keyframe motion latent $\boldsymbol{M}_k$. Subsequently, we iteratively denoise $\boldsymbol{M}_T$ using our trained denoising network, following the DDIM \cite{song2020denoising}. Finally, we obtain the predicted motion latent $\hat{\boldsymbol{M}}$, and utilizing the VAE decoder $\mathcal{H}$, we generate the RGB video $\hat{\boldsymbol{V}}=\mathcal{H}(\hat{\boldsymbol{M}}, \boldsymbol{I})$.
\section{Experiments}
\update{
We evaluate our model for both emotional talking control and facial motion control. For emotional talking control, the input includes audio and emotional guidance (such as labels, driving videos, and text, depending on the method). For facial motion control, we use only textual instructions to drive the avatar.
 }
\subsection{Experimental Setups}
\subsubsection{Datasets}

For emotional talking control, we augment the MEAD dataset~\cite{wang2020mead} following the methods outlined in~\cref{sec: method_data}. MEAD is a large-scale emotional talking face dataset featuring 8 emotion types and 3 intensity levels. We reserved 5 individuals for testing purposes and utilized the remaining data for training.  For text-guided facial motion control, we leveraged the CC v1 dataset~\cite{hazirbas2021towards}, which offers paired data comprising instructions and corresponding action videos. To ensure effective lip synchronization, we also incorporated the HDTF dataset~\cite{zhang2021flow}, which has high-quality talking face recordings. Our model was trained on a combination of these three datasets, enabling it to process either audio inputs or textual instructions. The evaluation was conducted using MEAD for in-domain assessment and TalkingHead 1KH~\cite{wang2021one} for out-of-domain evaluation.  
In the out-of-domain setting, the appearance information will be sourced from TalkingHead 1KH, while a randomly selected sample from the MEAD test set will serve as emotion guidance. \update{~\cref{tab:data_mod} in the~\cref{sec:app_mod} summarizes the available modalities and the corresponding tasks for each dataset.}
Further details about dataset statistics and preprocessing methods are also provided in the~\cref{sec:app_data}.

\subsubsection{Implementation Details} 

As detailed in~\cref{sec: method_arch}, we use Conformer~\cite{gulati2020conformer} as the backbone of our diffusion-based motion generator. Specifically, the model comprises 12 Conformer blocks, with a hidden state size of 768. For encoding textual instructions, we apply CLIP-L/14~\cite{CLIP}, and the Adapters are two layers MLPs. We adopt the Adam~\cite{kingma2014adam} optimizer and train our models on 8 V100 GPUs. The learning rate starts from $1e-5$ and follows the inverse square root schedule. More hyperparameters and architecture details can be found in the~\cref{sec:app_impl}.

\subsubsection{Evaluation Metrics}

\paragraph{\textbf{Objective Metrics}} \label{sec: obj_metrics}
To assess the effectiveness of emotion and motion control as well as the quality of talking head generation, we employ various evaluation metrics from different aspects.
To measure the fine-grained controlling ability, we propose AU$\rm{_{\textbf{F1}}}$, which calculates the F1 score of action units between the generated results and the ground truth. Furthermore, we introduce AU$\rm{_{\textbf{Emo}}}$, calculated as how many action units could be recalled by typical AUs of specific emotion types, to evaluate the overall coverage of facial details w.r.t an emotion type in the generated video. For motion control, we introduce the CLIP$\rm{_{\textbf{S}}}$ metric, which computes the CLIP embedding similarity between the text instruction and each frame, with the maximum value indicating the correspondence between the instruction and generated motion. Moreover, Sync$\rm{_{\textbf{D}}}$ is utilized to gauge lip-sync quality using SyncNet~\cite{chung2017out}, and FID~\cite{heusel2017gans} for both emotion and motion control to evaluate overall quality. Detailed definitions for some metrics can be found in~\cref{sec:app_metrics}.

\paragraph{\textbf{Subjective Metrics}}

We conduct the subjective evaluation involving 15 experienced users to score the generation quality and controllability of each model. Our evaluation metric is the Mean Opinion Score (MOS). For emotional talking control, we assess the lip-sync quality (Lip.), emotion controllability (Emo.), naturalness (Nat.), and motion jittering (Jit.). For facial motion control, we measured the accuracy of instruction following (Mot.) and identity preservation (ID.). Participants were presented with one video at a time and asked to rate each video for each score on a scale of 1 to 5. We calculated the average score as the final result.
\begin{table}[t]
\vspace{-0mm}
\tabcolsep=0.1cm
\begin{center}
\small
\caption{Quantitative comparison with baselines for \textbf{in-domain/out-of-the-domain} settings. The bold values indicate the best results, while the underlined values represent the second-best. Guid. Mod. indicates the modality of emotional guidance. Since there is no ground truth video in the out-of-the-domain setting, the FID metric is left empty. It can be observed that our model outperforms the baselines across many metrics. Notably, for Sync$\rm{_{\textbf{D}}}$, the ground truth video has a Sync$\rm{_{\textbf{D}}}$ of 9.172 in the in-domain setting, which is the closest to our model.}
\vspace{-0.0mm}
	\label{tab: baselines}
 \scalebox{0.93}{
	\begin{tabular}{lcccccc}
		\toprule[1.5pt]
		 Method  
         & {AU$\rm{_{\textbf{F1}}}$$\uparrow$} & {AU$\rm{_{\textbf{Emo}}}$$\uparrow$} & 
         {FID$\downarrow$} & {Sync$\rm{_{\textbf{D}}}$$\downarrow$} & Guid. Mod. \\
        \midrule
        
         GAIA~\cite{he2023gaia}                 & $0.549$/$0.185$ &  $0.352$/$0.048$ & $52.716$/$-$ & $9.542$/$9.776$ & - \\
        MakeItTalk~\cite{zhou2020makelttalk}                 & $0.588$/$0.220$ & $0.405$/$0.065$ & $\underline{47.269}$/$-$ & $11.291$/$10.059$ & -   \\

        EAT~\cite{eat}           & $0.648$/$\underline{0.542}$ & $0.495$/$\underline{0.319}$ & $57.379$/$-$ & $\bf{8.757}$/$\bf{8.962}$ & Label      \\
        
        StyleTalk~\cite{ma2023styletalk}        & $0.694$/$0.499$ & $\bf{0.593}$/$0.278$ & $75.783$/$-$ & $12.287$/$12.388$ &Video  \\
        
        DreamTalk~\cite{ma2023dreamtalk}        & $\underline{0.711}$/$0.513$ & $0.548$/$0.301$ & $85.291$/$-$ & $11.967$/$10.967$ & Video   \\
        
        \midrule
        \modelname~(Ours) & $\bf{0.738}$/$\bf{0.552}$ & $\underline{0.566}$/$\bf{0.324}$ & $\bf{44.593}$/$-$ & $\underline{9.412}$/$\underline{9.653}$ & Text \\
             
		\bottomrule[1.5pt]
	\end{tabular}
 }
\end{center}
\vspace{-1mm}
\end{table}

        

             
\begin{table}[t]
\vspace{-0mm}
\begin{center}
\small
\tabcolsep=0.1cm
\caption{Subjective evalution results for \modelname and other baselines for \textbf{in-domain/out-of-the-domain} settings. The bold values indicate the best results, while the underlined values represent
the second-best. 
}
\vspace{-0.0mm}
	\label{tab: subj}
	\begin{tabular}{lccccc}
		\toprule[1.5pt]
		 Method  
         & {Emo.$\uparrow$} &  {Lip.$\uparrow$} & {Jit.$\uparrow$} & 
         {Nat.$\uparrow$}   &\update{Guid. Mod.
}\\        \midrule
        
          GAIA~\cite{he2023gaia}                 & 3.02/1.83 &  \underline{4.63}/\underline{4.51} & \underline{4.57}/\underline{4.55} &  \underline{4.31}/\underline{4.47}&
         -
         \\
        MakeItTalk~\cite{zhou2020makelttalk}                 & 3.88/2.36 & 3.49/3.62 & 3.99/3.68 & 3.82/3.46
        &
         -\\

        EAT~\cite{eat}           & 4.48/\underline{4.18} & 4.27/4.18 & 4.40/3.27 & 4.14/4.20      &
         Label\\
        
        StyleTalk~\cite{ma2023styletalk}        & 4.53/4.12 & 4.17/2.26 & 3.43/1.88 & 3.18/2.04 &
         Video  \\
        
        DreamTalk~\cite{ma2023dreamtalk}        & \underline{4.59}/4.07 & 3.74/2.39 & 3.95/2.21 & 3.72/2.26  &
         Video  \\
        
        \midrule
        ~(Ours) & \textbf{4.64}/\textbf{4.52} & \textbf{4.74}/\textbf{4.59} & \textbf{4.88}/\textbf{4.68} & \textbf{4.63}/\textbf{4.60} &
         Text \\
             
		\bottomrule[1.5pt]
	\end{tabular}
\end{center}
\vspace{-1mm}
\end{table}

\begin{figure}
    \centering
    \includegraphics[width=1\textwidth]{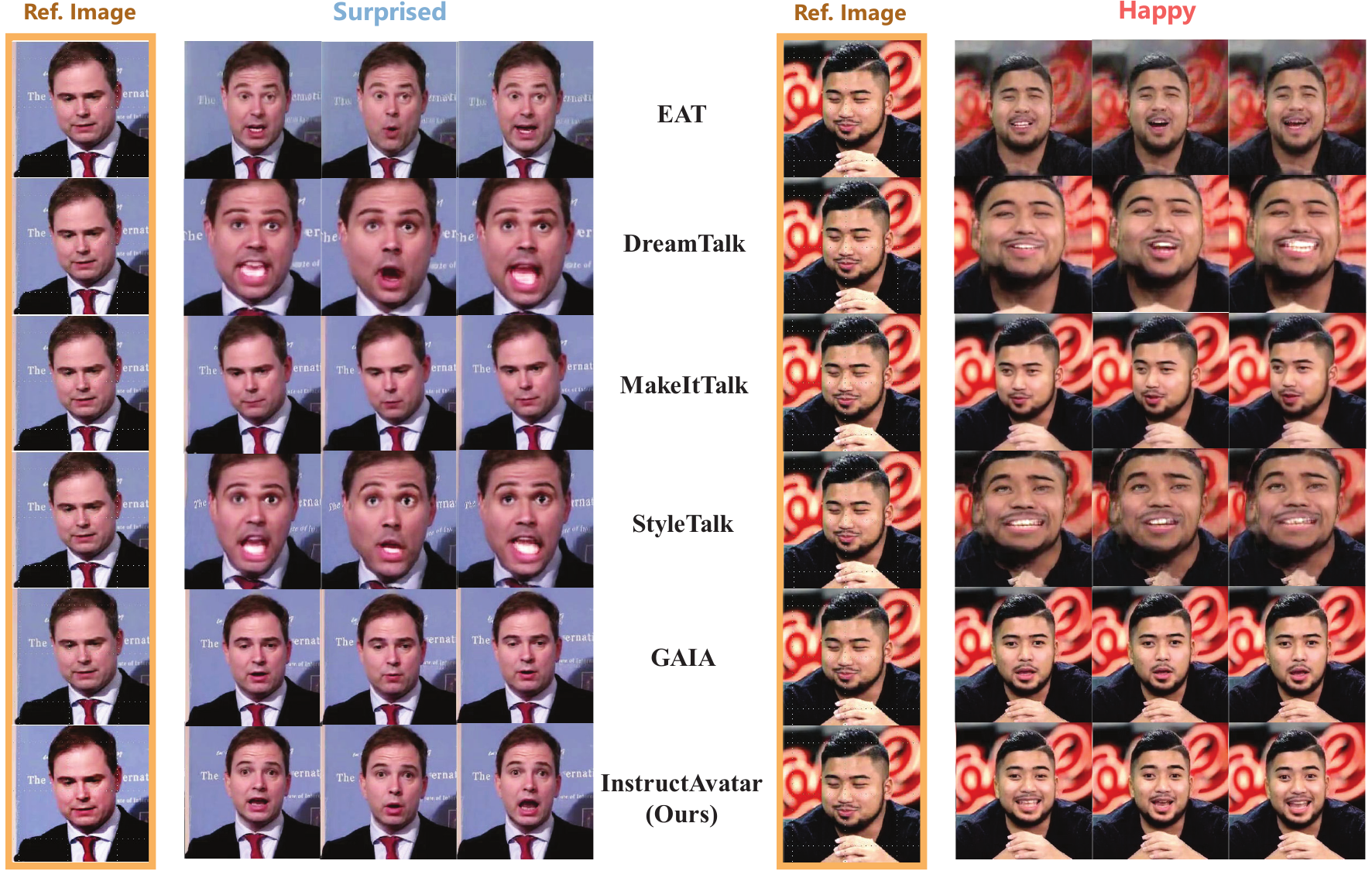}
    \caption{Qualitative comparison with baselines. It shows that \modelname achieves well lip-sync quality and emotion controllability. Additionally, the outputs generated by our model exhibit enhanced naturalness and effectively preserve identity characteristics.}
    \label{fig:baseline}
\end{figure}

\vspace{-0.3cm}
\subsection{Experimental Results}
\subsubsection{Emotional Talking Control}
\paragraph{\textbf{Baselines}}

We compare \modelname with several state-of-the-art methods. 
We consider emotion-unaware models GAIA \cite{he2023gaia} and MakeItTalk \cite{zhou2020makelttalk}, and provide the portrait and audio from the MEAD test set in inference. For the emotional label-based method EAT~\cite{eat}, we supplement the model with additional ground truth emotion types obtained in the annotation. Reference video-based methods like StyleTalk~\cite{ma2023styletalk} and DreamTalk \cite{ma2023dreamtalk} utilize the ground truth emotional video in MEAD as style guidance. It is worth noting that video-based methods utilize more information in generation than ours.

\paragraph{\textbf{Results}}

As shown in~\cref{tab: baselines}, the proposed \modelname model exhibits strong performance across most evaluation metrics, both in the in-domain and out-of-domain settings. Notably, our model demonstrates excellent fine-grained control ability, as reflected by AU$\rm{_{\textbf{F1}}}$ scores. Moreover, in the in-domain setting, our Sync$\rm{_{\textbf{D}}}$ metric is closer to the ground truth video (9.172) and exhibits better FID scores. It's worth mentioning that our model infers talking emotion solely based on text inputs, which intuitively poses a more challenging task. Additionally, our model supports a broader scope of instructions beyond high-level emotion types, which is absent for most baselines.

As a generative model, qualitative metrics may not fully capture the true effect. Therefore, we conducted subjective evaluation and case studies, as shown in~\cref{tab: subj} and~\cref{fig:baseline}, respectively. From these results, our model exhibits strong lip-sync ability and surpasses GAIA, a recently state-of-the-art audio-driven talking head model. Additionally, our model demonstrates superior emotion controllability compared to baselines tailored for emotional talking, such as EAT. More importantly, the generated results of our model appear more natural and robust to portrait images, as reflected by the Nat. metric and the cases shown in~\cref{fig:baseline}. More results are in~\cref{sec:app_emo}.

\subsubsection{Facial Motion Control}

\begin{figure}
    \centering
    \includegraphics[width=1\textwidth]{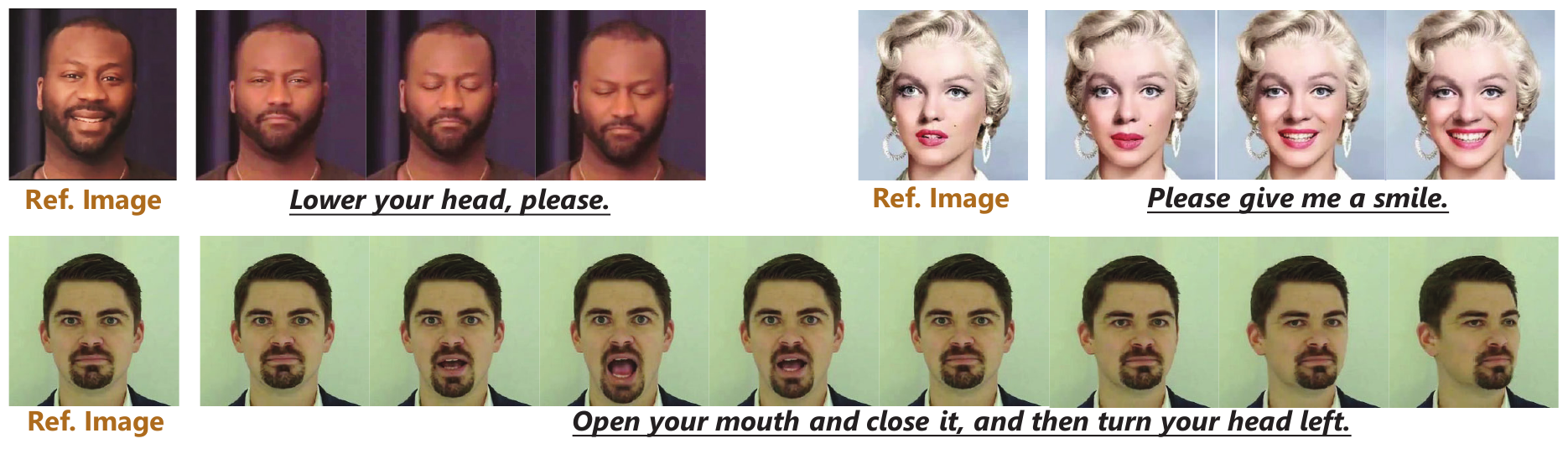}
    \caption{Examples of text-guided facial motion control.  Our model can execute precise motion control and generalize effectively to connect different actions.}
    \label{fig:t2m}
\end{figure}

To evaluate the effectiveness of the motion controllability, we establish four evaluation settings: \textbf{(\romannumeral1)} Simply repeating the portrait image (Ref.), \textbf{(\romannumeral2)} Given a random instruction (Rand. Inst.), \textbf{(\romannumeral3)} Given true instruction (GT Inst.), and \textbf{(\romannumeral4)} Ground-truth video (GT video). We set up \textbf{(\romannumeral1)} as a baseline to assess whether our model can generate dynamic video, and \textbf{(\romannumeral2)} to evaluate whether our model can follow instructions.  Results for different metrics in these settings are presented in~\cref{tab: t2m}, and visual results are provided in~\cref{fig:t2m}. Note that the GT Inst. setting represents the typical inference manner. Our model exhibits accurate instruction-following ability, as evidenced by similar CLIP$\rm{_{\textbf{S}}}$ metrics with the ground-truth video and a large performance gap with Rand. Inst., along with high subjective motion accuracy (Mot.) scores. Moreover, our model demonstrates excellent video generation quality, producing natural portraits while maintaining identity, as indicated by the ID. metric and \cref{fig:t2m}. Additionally, it generalizes effectively to connect different actions, as shown in~\cref{fig:t2m}. Please refer to~\cref{sec:app_motion} for more results.

\begin{table}[!th]
\centering
\vspace{0pt} 
    \begin{minipage}{0.49\linewidth}
        \centering
        \small
          \caption{Objective and subjective metrics for text-guided facial motion control.}
            \label{tab: t2m}
\begin{tabular}{lcccc}
        \toprule[1.5pt]
         \multirow{2}{*}{Methods}  & \multicolumn{2}{c}{Obj. Met.} & \multicolumn{2}{c}{Subj. Met.}\\
         \cmidrule(l){2-3}
          \cmidrule(l){4-5}
         & {FID$\downarrow$} & {CLIP$\rm{_{\textbf{S}}}$$\uparrow$} &{Mot.$\uparrow$}  & {ID.$\uparrow$}   \\
        \midrule
      \textcolor{gray}{Ref.} & \textcolor{gray}{$-$} & \textcolor{gray}{$20.500$} & \textcolor{gray}{$0$} & \textcolor{gray}{$-$}  \\
        \textcolor{gray}{Rand. Inst.} & \textcolor{gray}{$27.782$} & \textcolor{gray}{$21.274$} & \textcolor{gray}{$0.19$}  & \textcolor{gray}{$4.85$} \\
        GT Inst. & $25.370$ & $23.043$ & $4.65$ & $4.86$ \\
        \textcolor{gray}{GT Video} & \textcolor{gray}{$-$} & \textcolor{gray}{$23.136$} & \textcolor{gray}{$5$} & \textcolor{gray}{$5$} \\

        \bottomrule[1.5pt]
    \end{tabular}

    \end{minipage}
    \hfill
    \begin{minipage}{0.49\linewidth}
           \caption{Ablation studies on the proposed techniques.}
            \label{tab:ablation}
            \small
        \centering
	\begin{tabular}{l|ccc}
		\toprule[1.5pt]
	    Methods & Sync$\rm{_{\textbf{D}}}$$\downarrow$ & AU$\rm{_{\textbf{F1}}}$$\uparrow$ & Mot.$\uparrow$  \\
		\midrule
            \modelname     & $9.653$ & $0.552$ & $4.52$ \\
            \midrule
            (a) w/o AU  & $9.843$ & $0.435$ & $-$   \\
            (b) w/o Branch     & $9.994$ & $0.488$ & $4.10$  \\
            (c) w/o Zero conv.     & $12.832$ & $0.434$ & $3.62$  \\
		(d) w/o Rand. sub.   & $9.597$ & $0.339$ & $-$  \\
		\bottomrule[1.5pt]
	\end{tabular}
    \end{minipage}
\end{table}

\subsection{Ablation Study}

We conduct ablation studies to verify the effectiveness of each component of our model, as presented in~\cref{tab:ablation}. We utilize Sync$\rm{_{\textbf{D}}}$ to reflect lip-sync quality, AU$\rm{_{\textbf{F1}}}$ to gauge emotion controllability, and Mot. to assess motion controllability.

We can see that \textbf{(a)} In terms of data format, when no action units are provided during training, the model loses its ability to capture fine-grained details, resulting in a decrease in AU$\rm{_{\textbf{F1}}}$. \textbf{(b)} Combining emotion learning and motion learning into a single branch leads to a contraction to some extent, negatively impacting the performance of both controls, as evidenced by the AU$\rm{_{\textbf{F1}}}$ and Mot. metrics. \textbf{(c)} The Zero Convolution mechanism, detailed in~\cref{sec: method_zero}, is designed to stabilize training and harvest abundant knowledge from pretrained talking face models. Removing this component dramatically deteriorates lip-sync quality, and also influences emotion and motion control. \textbf{(d)} To prevent emotion leakage, we substitute the key frame latent with another emotion type during training emotional talking data. We found that this method significantly improves emotion control ability during out-of-domain tests, where portraits typically exhibit neutral emotion.
More ablation studies can be found in the~\cref{sec:app_ablation}.

\section{Conclusion}

In this paper, we introduce \modelname, a novel text-guided unified framework for emotion and motion control in avatar generation, significantly enhancing controllability and vividness compared to previous models. 
We develop an automatic annotation pipeline to construct a fine-grained and diverse instruction-video paired dataset and propose
a two-branch diffusion-based generator, with one branch focusing on emotion and the other on motion, to achieve fine-grained and accurate controls.
Experimental results demonstrate \modelname's exceptional lip-sync quality, fine-grained emotion controllability, user-friendly control interface, and naturalness of the generated outputs. We hope our work will inspire further research into text-guided emotional talking heads and anticipate more studies in this area. 
\paragraph{\textbf{Limitations}} Our work still has limitations. For example, our model is trained solely on a combination of action units extracted from real talking videos. This dependency between action units may limit its ability to precisely control a disentangled single action unit. Additionally, the relatively modest size of our training dataset may hinder its robustness when faced with highly out-of-domain instructions or appearances. \update{Moreover, since almost all data in our training dataset follows a single emotion/motion pattern, it is challenging for our model to control both emotion and motion simultaneously.}
We leave these challenges for future exploration.


%
%
\bibliographystyle{splncs04}
\bibliography{ref}

\begin{thebibliography}{10}
\providecommand{\url}[1]{\texttt{#1}}
\providecommand{\urlprefix}{URL }
\providecommand{\doi}[1]{https://doi.org/#1}

\bibitem{baevski2020wav2vec}
Baevski, A., Zhou, Y., Mohamed, A., Auli, M.: wav2vec 2.0: A framework for self-supervised learning of speech representations. Advances in neural information processing systems  \textbf{33},  12449--12460 (2020)

\bibitem{canfes2023text}
Canfes, Z., Atasoy, M.F., Dirik, A., Yanardag, P.: Text and image guided 3d avatar generation and manipulation. In: Proceedings of the IEEE/CVF Winter Conference on Applications of Computer Vision. pp. 4421--4431 (2023)

\bibitem{cao2014crema}
Cao, H., Cooper, D.G., Keutmann, M.K., Gur, R.C., Nenkova, A., Verma, R.: Crema-d: Crowd-sourced emotional multimodal actors dataset. IEEE transactions on affective computing  \textbf{5}(4),  377--390 (2014)

\bibitem{chu2024gpavatar}
Chu, X., Li, Y., Zeng, A., Yang, T., Lin, L., Liu, Y., Harada, T.: Gpavatar: Generalizable and precise head avatar from image(s) (2024)

\bibitem{chung2017out}
Chung, J.S., Zisserman, A.: Out of time: automated lip sync in the wild. In: Computer Vision--ACCV 2016 Workshops: ACCV 2016 International Workshops, Taipei, Taiwan, November 20-24, 2016, Revised Selected Papers, Part II 13. pp. 251--263. Springer (2017)

\bibitem{defossez2020real}
Defossez, A., Synnaeve, G., Adi, Y.: Real time speech enhancement in the waveform domain. arXiv preprint arXiv:2006.12847  (2020)

\bibitem{ekman1978facial}
Ekman, P., Friesen, W.V.: Facial action coding system. Environmental Psychology \& Nonverbal Behavior  (1978)

\bibitem{eat}
Gan, Y., Yang, Z., Yue, X., Sun, L., Yang, Y.: Efficient emotional adaptation for audio-driven talking-head generation. In: Proceedings of the IEEE/CVF International Conference on Computer Vision. pp. 22634--22645 (2023)

\bibitem{gao2024clip}
Gao, P., Geng, S., Zhang, R., Ma, T., Fang, R., Zhang, Y., Li, H., Qiao, Y.: Clip-adapter: Better vision-language models with feature adapters. International Journal of Computer Vision  \textbf{132}(2),  581--595 (2024)

\bibitem{goodfellow2014generative}
Goodfellow, I., Pouget-Abadie, J., Mirza, M., Xu, B., Warde-Farley, D., Ozair, S., Courville, A., Bengio, Y.: Generative adversarial nets. Advances in neural information processing systems  \textbf{27} (2014)

\bibitem{gulati2020conformer}
Gulati, A., Qin, J., Chiu, C.C., Parmar, N., Zhang, Y., Yu, J., Han, W., Wang, S., Zhang, Z., Wu, Y., et~al.: Conformer: Convolution-augmented transformer for speech recognition. arXiv preprint arXiv:2005.08100  (2020)

\bibitem{gururani2023space}
Gururani, S., Mallya, A., Wang, T.C., Valle, R., Liu, M.Y.: Space: Speech-driven portrait animation with controllable expression. In: Proceedings of the IEEE/CVF International Conference on Computer Vision. pp. 20914--20923 (2023)

\bibitem{hazirbas2021towards}
Hazirbas, C., Bitton, J., Dolhansky, B., Pan, J., Gordo, A., Ferrer, C.C.: Towards measuring fairness in ai: the casual conversations dataset. IEEE Transactions on Biometrics, Behavior, and Identity Science  \textbf{4}(3),  324--332 (2021)

\bibitem{he2023gaia}
He, T., Guo, J., Yu, R., Wang, Y., Zhu, J., An, K., Li, L., Tan, X., Wang, C., Hu, H., et~al.: Gaia: Zero-shot talking avatar generation. ICLR  (2024)

\bibitem{heusel2017gans}
Heusel, M., Ramsauer, H., Unterthiner, T., Nessler, B., Hochreiter, S.: Gans trained by a two time-scale update rule converge to a local nash equilibrium. Advances in neural information processing systems  \textbf{30} (2017)

\bibitem{ho2020denoising}
Ho, J., Jain, A., Abbeel, P.: Denoising diffusion probabilistic models. Advances in neural information processing systems  \textbf{33},  6840--6851 (2020)

\bibitem{huang2023collaborative}
Huang, Z., Chan, K.C., Jiang, Y., Liu, Z.: Collaborative diffusion for multi-modal face generation and editing. In: Proceedings of the IEEE/CVF Conference on Computer Vision and Pattern Recognition. pp. 6080--6090 (2023)

\bibitem{ji2022eamm}
Ji, X., Zhou, H., Wang, K., Wu, Q., Wu, W., Xu, F., Cao, X.: Eamm: One-shot emotional talking face via audio-based emotion-aware motion model. In: ACM SIGGRAPH 2022 Conference Proceedings. pp. 1--10 (2022)

\bibitem{kingma2014adam}
Kingma, D.P., Ba, J.: Adam: A method for stochastic optimization. arXiv preprint arXiv:1412.6980  (2014)

\bibitem{li2023instruct}
Li, S.: Instruct-video2avatar: Video-to-avatar generation with instructions. arXiv preprint arXiv:2306.02903  (2023)

\bibitem{liu2024towards}
Liu, R., Ma, B., Zhang, W., Hu, Z., Fan, C., Lv, T., Ding, Y., Cheng, X.: Towards a simultaneous and granular identity-expression control in personalized face generation. arXiv preprint arXiv:2401.01207  (2024)

\bibitem{liu2022audio}
Liu, X., Wu, Q., Zhou, H., Du, Y., Wu, W., Lin, D., Liu, Z.: Audio-driven co-speech gesture video generation. Advances in Neural Information Processing Systems  \textbf{35},  21386--21399 (2022)

\bibitem{audetect}
Luo, C., Song, S., Xie, W., Shen, L., Gunes, H.: Learning multi-dimensional edge feature-based au relation graph for facial action unit recognition. In: Proceedings of the Thirty-First International Joint Conference on Artificial Intelligence. IJCAI-2022, International Joint Conferences on Artificial Intelligence Organization (Jul 2022). \doi{10.24963/ijcai.2022/173}, \url{http://dx.doi.org/10.24963/ijcai.2022/173}

\bibitem{ma2023talkclip}
Ma, Y., Wang, S., Ding, Y., Ma, B., Lv, T., Fan, C., Hu, Z., Deng, Z., Yu, X.: Talkclip: Talking head generation with text-guided expressive speaking styles. arXiv preprint arXiv:2304.00334  (2023)

\bibitem{ma2023styletalk}
Ma, Y., Wang, S., Hu, Z., Fan, C., Lv, T., Ding, Y., Deng, Z., Yu, X.: Styletalk: One-shot talking head generation with controllable speaking styles. arXiv preprint arXiv:2301.01081  (2023)

\bibitem{ma2023dreamtalk}
Ma, Y., Zhang, S., Wang, J., Wang, X., Zhang, Y., Deng, Z.: Dreamtalk: When expressive talking head generation meets diffusion probabilistic models. arXiv preprint arXiv:2312.09767  (2023)

\bibitem{gpt4v}
OpenAI: Gpt-4v(ision) system card  (2023)

\bibitem{patashnik2021styleclip}
Patashnik, O., Wu, Z., Shechtman, E., Cohen-Or, D., Lischinski, D.: Styleclip: Text-driven manipulation of stylegan imagery. In: Proceedings of the IEEE/CVF International Conference on Computer Vision. pp. 2085--2094 (2021)

\bibitem{CLIP}
Radford, A., Kim, J.W., Hallacy, C., Ramesh, A., Goh, G., Agarwal, S., Sastry, G., Askell, A., Mishkin, P., Clark, J., et~al.: Learning transferable visual models from natural language supervision. In: International conference on machine learning. pp. 8748--8763. PMLR (2021)

\bibitem{ren2021pirenderer}
Ren, Y., Li, G., Chen, Y., Li, T.H., Liu, S.: Pirenderer: Controllable portrait image generation via semantic neural rendering. In: Proceedings of the IEEE/CVF International Conference on Computer Vision. pp. 13759--13768 (2021)

\bibitem{siarohin2019first}
Siarohin, A., Lathuili{\`e}re, S., Tulyakov, S., Ricci, E., Sebe, N.: First order motion model for image animation. Advances in neural information processing systems  \textbf{32} (2019)

\bibitem{song2020denoising}
Song, J., Meng, C., Ermon, S.: Denoising diffusion implicit models. arXiv preprint arXiv:2010.02502  (2020)

\bibitem{sun2023vividtalk}
Sun, X., Zhang, L., Zhu, H., Zhang, P., Zhang, B., Ji, X., Zhou, K., Gao, D., Bo, L., Cao, X.: Vividtalk: One-shot audio-driven talking head generation based on 3d hybrid prior. arXiv preprint arXiv:2312.01841  (2023)

\bibitem{emmn}
Tan, S., Ji, B., Pan, Y.: Emmn: Emotional motion memory network for audio-driven emotional talking face generation. pp. 22089--22099 (10 2023). \doi{10.1109/ICCV51070.2023.02024}

\bibitem{tian2024emo}
Tian, L., Wang, Q., Zhang, B., Bo, L.: Emo: Emote portrait alive - generating expressive portrait videos with audio2video diffusion model under weak conditions (2024)

\bibitem{tripathy2021facegan}
Tripathy, S., Kannala, J., Rahtu, E.: Facegan: Facial attribute controllable reenactment gan. In: Proceedings of the IEEE/CVF winter conference on applications of computer vision. pp. 1329--1338 (2021)

\bibitem{pd-fgc}
Wang, D., Deng, Y., Yin, Z., Shum, H.Y., Wang, B.: Progressive disentangled representation learning for fine-grained controllable talking head synthesis. In: Proceedings of the IEEE/CVF Conference on Computer Vision and Pattern Recognition. pp. 17979--17989 (2023)

\bibitem{wang2024facecomposer}
Wang, J., Zhao, K., Ma, Y., Zhang, S., Zhang, Y., Shen, Y., Zhao, D., Zhou, J.: Facecomposer: A unified model for versatile facial content creation. Advances in Neural Information Processing Systems  \textbf{36} (2024)

\bibitem{wang2020mead}
Wang, K., Wu, Q., Song, L., Yang, Z., Wu, W., Qian, C., He, R., Qiao, Y., Loy, C.C.: Mead: A large-scale audio-visual dataset for emotional talking-face generation. In: European Conference on Computer Vision. pp. 700--717. Springer (2020)

\bibitem{wang2021audio2head}
Wang, S., Li, L., Ding, Y., Fan, C., Yu, X.: Audio2head: Audio-driven one-shot talking-head generation with natural head motion. arXiv preprint arXiv:2107.09293  (2021)

\bibitem{wang2022one}
Wang, S., Li, L., Ding, Y., Yu, X.: One-shot talking face generation from single-speaker audio-visual correlation learning. In: Proceedings of the AAAI Conference on Artificial Intelligence. vol.~36, pp. 2531--2539 (2022)

\bibitem{wang2021facevid2vid}
Wang, T.C., Mallya, A., Liu, M.Y.: One-shot free-view neural talking-head synthesis for video conferencing. In: Proceedings of the IEEE Conference on Computer Vision and Pattern Recognition (2021)

\bibitem{wang2021one}
Wang, T.C., Mallya, A., Liu, M.Y.: One-shot free-view neural talking-head synthesis for video conferencing. In: Proceedings of the IEEE/CVF conference on computer vision and pattern recognition. pp. 10039--10049 (2021)

\bibitem{xia2021tedigan}
Xia, W., Yang, Y., Xue, J.H., Wu, B.: Tedigan: Text-guided diverse face image generation and manipulation. In: Proceedings of the IEEE/CVF conference on computer vision and pattern recognition. pp. 2256--2265 (2021)

\bibitem{xu2023high}
Xu, C., Zhu, J., Zhang, J., Han, Y., Chu, W., Tai, Y., Wang, C., Xie, Z., Liu, Y.: High-fidelity generalized emotional talking face generation with multi-modal emotion space learning. In: Proceedings of the IEEE/CVF Conference on Computer Vision and Pattern Recognition. pp. 6609--6619 (2023)

\bibitem{yu2023towards}
Yu, C., Lu, G., Zeng, Y., Sun, J., Liang, X., Li, H., Xu, Z., Xu, S., Zhang, W., Xu, H.: Towards high-fidelity text-guided 3d face generation and manipulation using only images. In: Proceedings of the IEEE/CVF International Conference on Computer Vision. pp. 15326--15337 (2023)

\bibitem{zhai2023talking}
Zhai, S., Liu, M., Li, Y., Gao, Z., Zhu, L., Nie, L.: Talking face generation with audio-deduced emotional landmarks. IEEE Transactions on Neural Networks and Learning Systems  (2023)

\bibitem{zhang2022metaportrait}
Zhang, B., Qi, C., Zhang, P., Zhang, B., Wu, H., Chen, D., Chen, Q., Wang, Y., Wen, F.: Metaportrait: Identity-preserving talking head generation with fast personalized adaptation. arXiv:2212.08062  (2022)

\bibitem{zhang2023dream}
Zhang, C., Wang, C., Zhang, J., Xu, H., Song, G., Xie, Y., Luo, L., Tian, Y., Guo, X., Feng, J.: Dream-talk: Diffusion-based realistic emotional audio-driven method for single image talking face generation. arXiv preprint arXiv:2312.13578  (2023)

\bibitem{zhang2023dreamface}
Zhang, L., Qiu, Q., Lin, H., Zhang, Q., Shi, C., Yang, W., Shi, Y., Yang, S., Xu, L., Yu, J.: Dreamface: Progressive generation of animatable 3d faces under text guidance. arXiv preprint arXiv:2304.03117  (2023)

\bibitem{zhang2023adding}
Zhang, L., Rao, A., Agrawala, M.: Adding conditional control to text-to-image diffusion models. In: Proceedings of the IEEE/CVF International Conference on Computer Vision. pp. 3836--3847 (2023)

\bibitem{zhang2023sadtalker}
Zhang, W., Cun, X., Wang, X., Zhang, Y., Shen, X., Guo, Y., Shan, Y., Wang, F.: Sadtalker: Learning realistic 3d motion coefficients for stylized audio-driven single image talking face animation. In: Proceedings of the IEEE/CVF Conference on Computer Vision and Pattern Recognition. pp. 8652--8661 (2023)

\bibitem{zhang2021flow}
Zhang, Z., Li, L., Ding, Y., Fan, C.: Flow-guided one-shot talking face generation with a high-resolution audio-visual dataset. In: Proceedings of the IEEE/CVF Conference on Computer Vision and Pattern Recognition. pp. 3661--3670 (2021)

\bibitem{zhao2024media2face}
Zhao, Q., Long, P., Zhang, Q., Qin, D., Liang, H., Zhang, L., Zhang, Y., Yu, J., Xu, L.: Media2face: Co-speech facial animation generation with multi-modality guidance. arXiv preprint arXiv:2401.15687  (2024)

\bibitem{zhong2023expclip}
Zhong, Y., Wei, H., Yang, P., Wang, Z.: Expclip: Bridging text and facial expressions via semantic alignment. arXiv preprint arXiv:2308.14448  (2023)

\bibitem{zhou2020makelttalk}
Zhou, Y., Han, X., Shechtman, E., Echevarria, J., Kalogerakis, E., Li, D.: Makelttalk: speaker-aware talking-head animation. ACM Transactions On Graphics (TOG)  \textbf{39}(6),  1--15 (2020)

\end{thebibliography}

\newpage
\appendix

\section{More Experimental Results}
\subsection{More Results about Emotional Talking}\label{sec:app_emo}

\begin{figure}
    \centering
\includegraphics[width=0.9\textwidth]{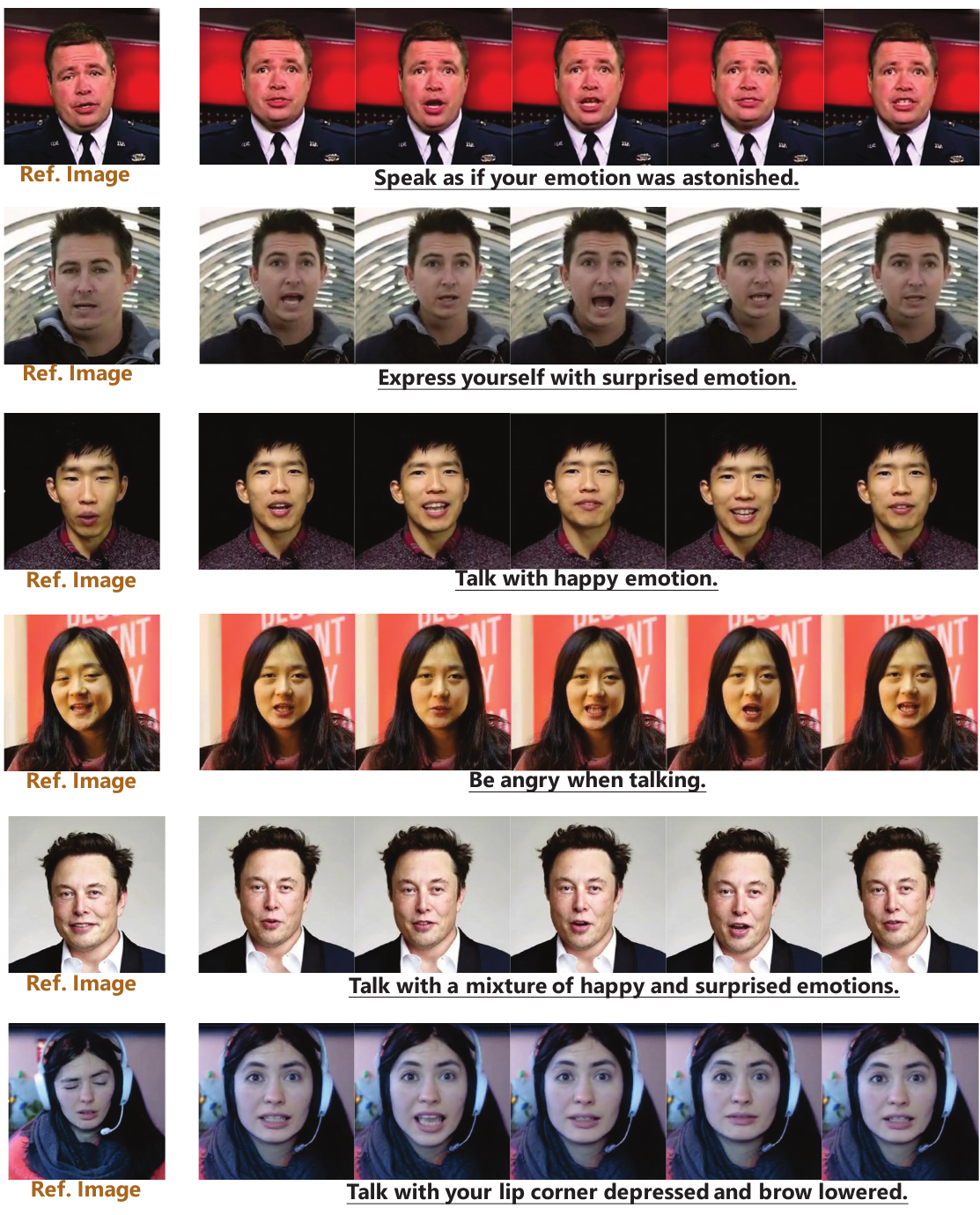}
    \caption{More examples of text-guided emotional talking control. }
    \label{fig:emo_cont}
\end{figure}

Additional outcomes concerning text-guided emotional talking control are presented in~\cref{fig:emo_cont}. We can see that our model exhibits precise emotion control ability, with the generated results appearing natural. Furthermore, \modelname supports fine-grained control and demonstrates reasonable generalization ability beyond the domain.

\begin{figure}
    \centering
    \includegraphics[width=1\textwidth]{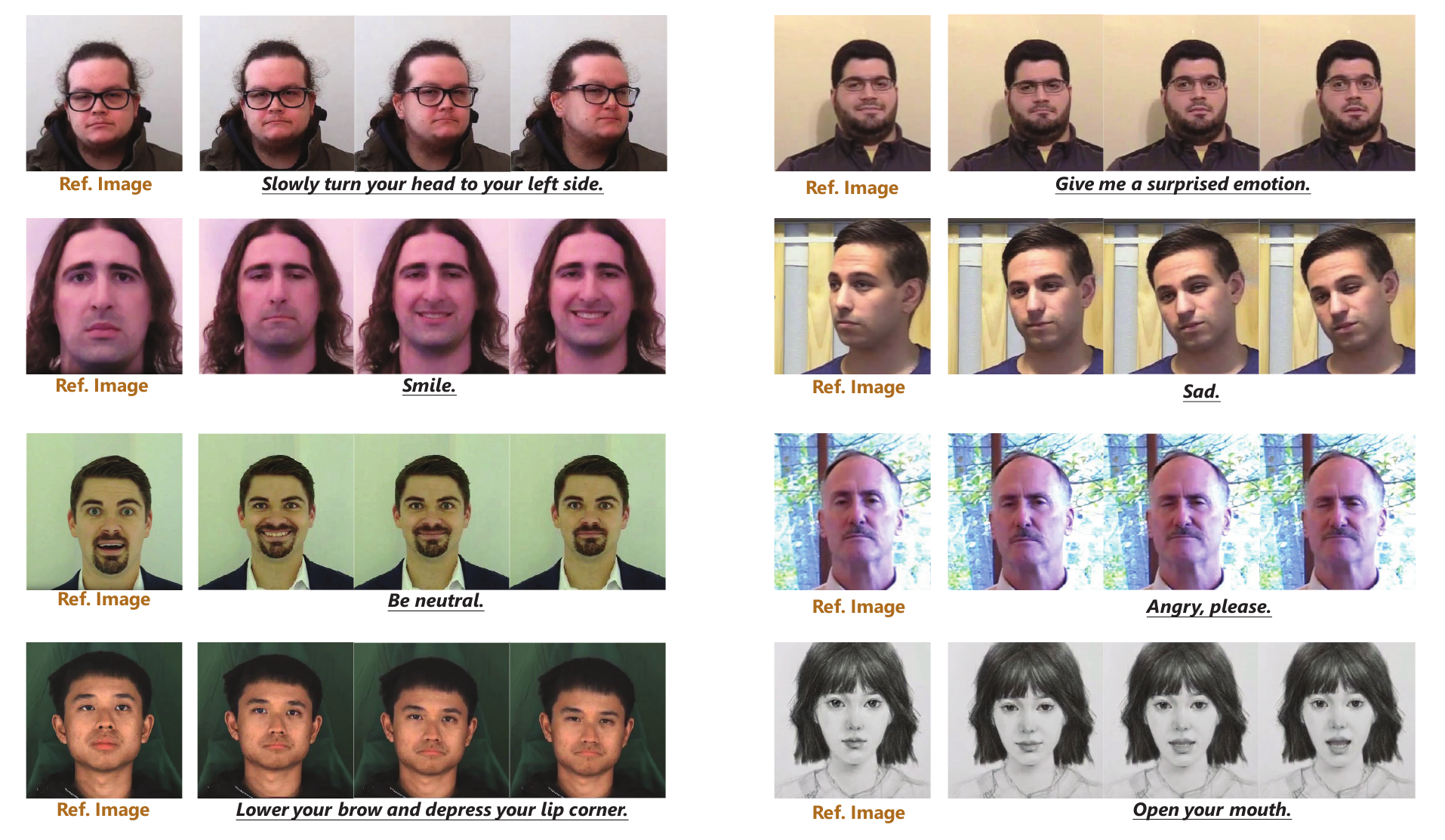}
    \caption{More examples of text-guided facial motion control. }
    \label{fig:t2m_cont}
\end{figure}
\subsection{More Results about Facial Motion Control}\label{sec:app_motion}

We show more results about facial motion control in \cref{fig:t2m_cont}. It is evident that \modelname exhibits remarkable proficiency in following instructions and preserving identity. Furthermore, the generated results appear natural and robust with variations in the provided portrait, including tilting or inherent expressions. Moreover, our model demonstrates fine-grained control capability and performs effectively in out-of-domain scenarios, as depicted in the last row of~\cref{fig:t2m_cont}.

\subsection{The Effectiveness of Textual Instructions}
\begin{figure}
    \centering
\includegraphics[width=0.9\textwidth]{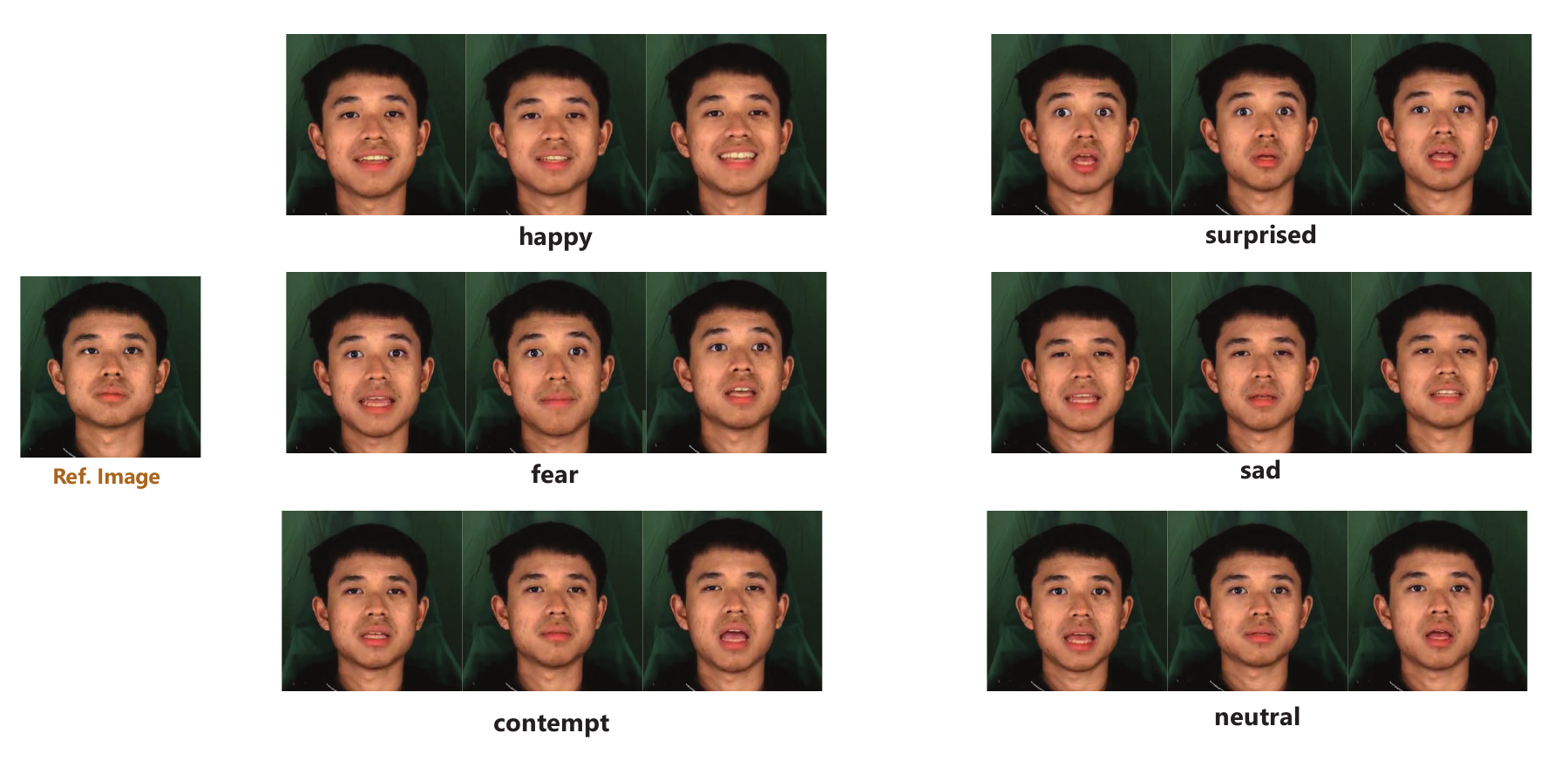}
    \caption{Illustration of the effectiveness of textual instructions.  All videos are generated utilizing identical portraits and neutral audio, with variations only in the textual instructions.}
    \label{fig:diff_emo}
\end{figure}

To animate the avatar in our model, we input three conditions: portrait, audio, and textual instructions. We acknowledge that in real life, all these conditions can convey emotion. Therefore, a natural question arises: Does the emotion depicted in our generated videos primarily stem from the textual instructions rather than from the portraits or the inherent emotion conveyed in the audio? We address this question in \cref{fig:diff_emo}, where all videos are generated using identical neutral portraits and neutral audio. The results demonstrate that our model can still produce distinct emotional talking videos, highlighting the effectiveness of textual instructions.
\subsection{Emotion Intensity}

\begin{figure}
    \centering
\includegraphics[width=0.9\textwidth]{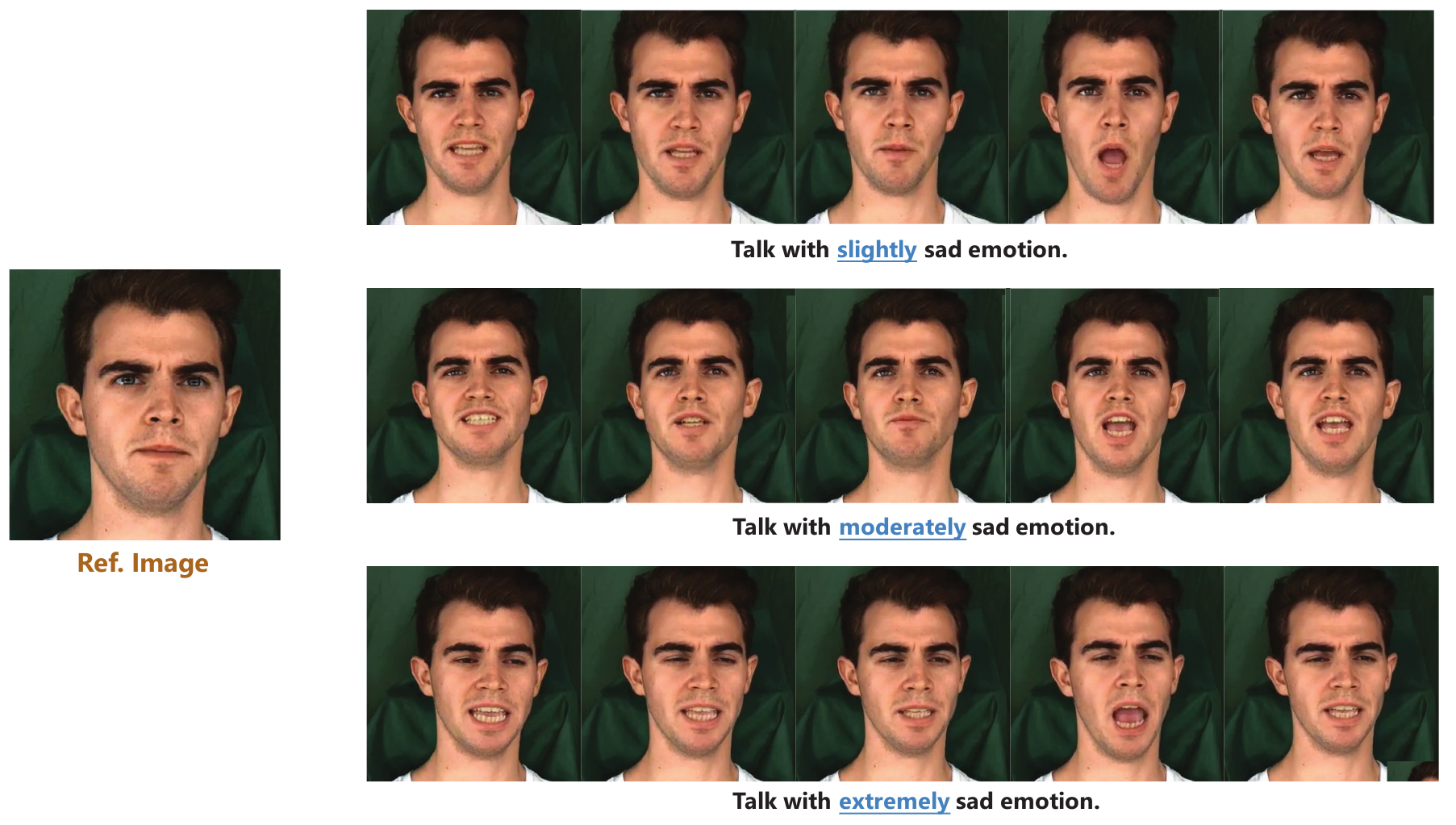}
    \caption{Illustration of emotion intensity control. }
    \label{fig:inten}
\end{figure}

\modelname demonstrates the capability to generate results with varying levels of emotion intensity. We illustrate a case in~\cref{fig:inten}. It is evident that our model distinguishes between different emotion intensities based on specific descriptors such as ``extremely'' and ``slightly''.

\begin{table}[t]
\begin{center}
\small
\caption{More ablation studies on the proposed techniques.}
\vspace{-0mm}
	\label{tab:ablation_cont}
	\begin{tabular}{l|ccc}
		\toprule[1.5pt]
	    Methods & Sync$\rm{_{\textbf{D}}}$$\downarrow$ & AU$\rm{_{\textbf{F1}}}$$\uparrow$ & Mot.$\uparrow$  \\
		\midrule
            \modelname     & $9.747$ & $0.537$ & $4.46$ \\
            \midrule
        (a) w/o CLIP Adapter   & $9.903$ & $0.512$ & $4.31$  \\
  	 (b) w/o AU loss   & $9.753$ & $0.504$ & $-$  \\
      
          	 (c) w/o Empty noise   & $10.427$ & $0.469$ & $3.23$  \\
	    (d) w/ Label input   & $9.763$ & $0.469$ & $-$   \\
            (e) All tokens for Emo.     & $9.842$ & $0.417$ & $-$  \\
            (f) \texttt{[EOS]} token for Mot.    & $-$ & $-$ & $4.03$ \\
            
		\bottomrule[1.5pt]
	\end{tabular}
\end{center}
\vspace{-1mm}
\end{table}

\subsection{More Ablation Results}\label{sec:app_ablation}
We provide more ablation results in~\cref{tab:ablation_cont}. To reduce the training cost, we randomly select 40\% of the samples from the training pool. Given the substantial number of samples (over 60k), the statistical distribution differences would be marginal, thus representing the entire dataset. We observe that:
\textbf{(a)} When the CLIP Adapter~\cite{gao2024clip} is removed, the model loses a converter from the CLIP text space to the space required by the denoising block. This results in a decrease in both AU$\rm{_{\textbf{F1}}}$ and Mot. metrics.
\textbf{(b)} When the AU loss is removed, the model loses some strict guidance on capturing fine-grained action unit details, leading to a decrease in AU$\rm{_{\textbf{F1}}}$. 
\textbf{(c)} 
To integrate facial motion control where no audio is provided into our unified framework, we use pseudo-empty audio as a placeholder. However, upon switching this strategy to employ another pseudo audio feature, such as tensors with all 0s, we observe a rapid deterioration in facial motion control, which also impacts emotional control, as indicated by Mot. and AU$\rm{_{\textbf{F1}}}$. We attribute this to the inherent physical meaning carried by pseudo-empty audio, symbolizing silence and resulting in a still avatar. Conversely, the use of fake audio features like tensors with all 0s lacks meaningful interpretation. Consequently, when combined with normal audio during training, the model becomes confused due to this misaligned setting.
\update{\textbf{(d)} Beyond the limited emotion categories used in previous models, we adopt natural language for an open-vocabulary emotion guidance approach. This method has several advantages as discussed in \cref{sec:intro}: (1) Enhanced control over fine-grained details rather than just the overall style; (2) Improved generalizability compared to limited emotion categories; and (3) Enhanced interactivity and user-friendliness. Experimentally, when training our model with emotion labels, as shown in \cref{tab:ablation_cont} (d), the analysis of AU$\rm{_{\textbf{F1}}}$ indicates that text guidance offers better fine-grained controllability. \textbf{(e)} The \texttt{[EOS]} token is well-suited for the emotional talking task, acting as a general style guide. Replacing it with all instruction tokens leads to a performance drop, demonstrating that additional tokens may distract the model from extracting overall information. \textbf{(f)} Conversely, the hidden states of all tokens are more suitable for the motion control task, which requires temporally dynamic guidance. Replacing these with only \texttt{[EOS]} tokens fails to reflect dynamic information, resulting in suboptimal performance.
}

\subsection{The Reliability of GPT-4V}\label{sec:app_gpt}

\begin{figure}
    \centering
\includegraphics[width=0.9\textwidth]{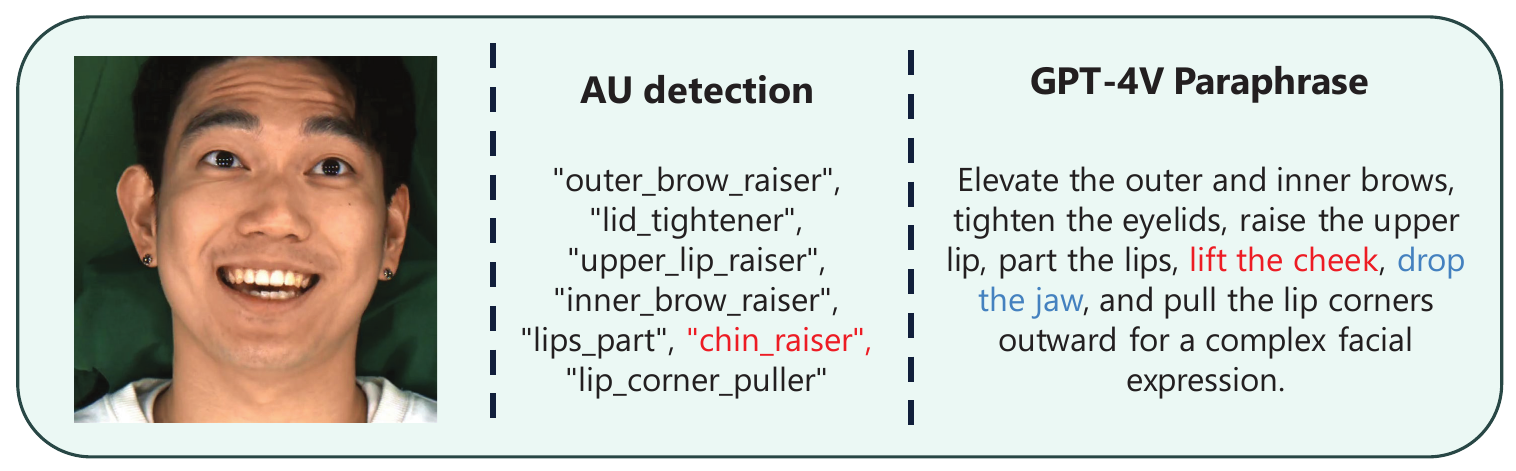}
    \caption{Illustration of the effectiveness of GPT-4V paraphrase. We can see that GPT-4V generates a fluent sentence while rectifying incorrect action units and supplementing omitted ones.}
    \label{fig:para}
\end{figure}

In~\cref{sec: method_data} in the main text, we mentioned that we would use GPT-4V~\cite{gpt4v} to transform the detected action units into coherent sentences and to reexamine the action units detected by off-the-shelf models, leveraging its visual capabilities. We provide an example wherein GPT-4V converts action units into a sentence, accurately rectifying erroneous action units and supplementing omitted ones, as depicted in \cref{fig:para}. Furthermore, we randomly examined 20 examples from the dataset and observed that in 16 instances, there was no significant difference in the action units detected with or without leveraging visual capabilities. In the remaining 4 cases, we favored the results obtained by GPT-4V, and in none of these cases did GPT-4V generate an incorrect response.

\update{
We also conduct a more extensive user study to evaluate the reliability of LLMs. We enlist 10 experienced volunteers to assess the outputs. For the 60 templates generated by ChatGPT, the volunteers found all to be satisfactory. For the paraphrased sentences generated by GPT-4V, we randomly select 30 instances (each with 3 sentences) and ask the volunteers to evaluate their accuracy, fluency, conciseness, diversity, and the correctness of overriding action units generated by the off-the-shelf model, rating them from 0 to 5. The results, shown in \cref{tab:gpt_eval}, indicate that LLMs not only help us quickly gather extensive training data but are also reliable.
}
\begin{table}[t]
\begin{center}
\small
\caption{Human evaluation for the application of GPT-4V on the data construction.}
\vspace{-0mm}
	\label{tab:gpt_eval}
	\begin{tabular}{l|c|c|c|c|c}
		\toprule[1.5pt]
	     & Accuracy  &Fluency &Conciseness& Diversity&Correctness of overriding\\
		\midrule
            \textbf{Average Score} & 4.7 & 4.9 & 4.7 & 4.5 & 4.8 \\
		\bottomrule[1.5pt]
	\end{tabular}
\end{center}
\vspace{-1mm}
\end{table}
\section{Metrics Definition}\label{sec:app_metrics}
We provide formal definitions for self-defined metrics in our paper.

\subsection{AU$\rm{_{\textbf{F1}}}$ and AU$\rm{_{\textbf{Emo}}}$}
To evaluate the fine-grained controllability of \modelname, we introduce AU$\rm{_{\textbf{F1}}}$ and AU$\rm{_{\textbf{Emo}}}$. Let us consider action units extracted from generated sample $j$, denoted by $\hat{\boldsymbol{y}}^{(j)} \in \mathbb{R}^M$, alongside corresponding ground truth action units $\boldsymbol{y}^{(j)} \in \mathbb{R}^M$, where $M$ represents the number of action units (in our paper $M = 41$). Both $\hat{\boldsymbol{y}}^{(j)}$ and $\boldsymbol{y}^{(j)}$ are vectors composed of 0s and 1s, with 1 indicating activation of the action unit and 0 indicating its inactivation.

To quantify the concordance between the action units of generated results and ground truth, assuming we have $n$ samples, we compute the F1 score for this multi-label classification problem as follows:
\[\text{AU$\rm{_{\textbf{F1}}}$} =\frac{1}{n}\sum_{j=1}^n \frac{2|\boldsymbol{y}^{(j)}\cap \hat{\boldsymbol{y}}^{(j)}|}{|\boldsymbol{y}^{(j)}| + |\hat{\boldsymbol{y}}^{(j)}|}=\frac{1}{n}\sum_{j=1}^n\frac{2\sum_{i=1}^M \boldsymbol{y}_i^{(j)}\cdot \hat{\boldsymbol{y}}_i^{(j)}}{\sum_{i=1}^M \boldsymbol{y}_i^{(j)} + \sum_{i=1}^M \hat{\boldsymbol{y}}_i^{(j)}}\]
Moreover, to evaluate the overall coverage of facial details with respect to an emotion type, we define AU$\rm{_{\textbf{Emo}}}$. Firstly, we identify typical and representative action unit combinations for each emotion, as shown in \cref{tab:class_au}. Then, for each generated result $\hat{\boldsymbol{y}}^{(j)}$, suppose the corresponding action units for such desired emotion are $\boldsymbol{y}_{emo}^{(j)}$, we calculate how many action units could be recalled by the typical action units:
\[\text{AU$\rm{_{\textbf{Emo}}}$} =\frac{1}{n}\sum_{j=1}^n \frac{2|\boldsymbol{y}_{emo}^{(j)}\cap \hat{\boldsymbol{y}}^{(j)}|}{|\boldsymbol{y}_{emo}^{(j)}|}=\frac{1}{n}\sum_{j=1}^n\frac{2\sum_{i=1}^M \boldsymbol{y}_{emo,i}^{(j)}\cdot \hat{\boldsymbol{y}}_i^{(j)}}{\sum_{i=1}^M \boldsymbol{y}_{emo,i}^{(j)} }\]

\subsection{CLIP$\rm{_{\textbf{S}}}$} We employ CLIP$\rm{_{\textbf{S}}}$ to measure the accuracy of text-guided motion control, as shown in~\cref{sec: obj_metrics} in the main text. Let $t$ denote the text instruction and $v$ denote the generated video. We denote $v_i$ as the $i$-th frame of video $v$. We use the CLIP~\cite{CLIP} text encoder $\mathcal{E}_t$ to encode the text and the image encoder $\mathcal{E}_v$ to encode each frame. Since the CLIP model is trained on paired image-text data, it possesses powerful modality alignment ability. We utilize cosine similarity to calculate the matchness of each frame with the instruction. Considering that motion control is a dynamic process, we consider it successful if it matches the instructions for a period of time. Therefore, we take the maximum of the similarity scores $s_i$ as our final result:
\[s = \mathop{\max}_{i} s_i = \mathop{\max}_{i} \frac{\mathcal{E}_t(t)\cdot \mathcal{E}_v(v_i)}{\|\mathcal{E}_t(t)\|\cdot \|\mathcal{E}_v(v_i)\|}\]

\begin{table}[t]
\begin{center}
\small
\caption{Typical Action Units for Different Emotions.}
\vspace{-0mm}
	\label{tab:class_au}
	\begin{tabular}{l|c}
		\toprule[1.5pt]
	    \textbf{Emotion Type} & \textbf{Typical Action Units}  \\
		\midrule
            Angry & Brow Lowerer, Jaw Drop, Nose Wrinkler, Lid Tightener \\
            \midrule
Fear & Inner Brow Raiser, Jaw Drop, Upper Lid Raiser, Outer Brow Raiser \\
\midrule
Happy & Cheek Raiser, Lip Corner Puller, Jaw Drop, Lid Tightener \\ \midrule
Contempt & Inner Brow Raiser, Chin Raiser, Lip Corner Puller, Lid Tightener \\ \midrule
Disgusted & Brow Lowerer, Cheek Raiser, Lip Corner Depressor, Nose Wrinkler \\ \midrule
Sad & Brow Lowerer, Chin Raiser, Inner Brow Raiser, Lip Corner Depressor \\
\midrule
Surprised & Inner Brow Raiser, Jaw Drop, Outer Brow Raiser, Lid Tightener \\
		\bottomrule[1.5pt]
	\end{tabular}
\end{center}
\vspace{-1mm}
\end{table}

\section{Data}\label{sec:app_data}
\update{
\subsection{Modalities of Each Dataset}\label{sec:app_mod}
In~\cref{tab:data_mod}, We list the available modalities and corresponding tasks of each dataset utilized in our paper.
\begin{table}
\centering 
\caption{Available modalities and corresponding tasks of the collected dataset. * indicates that for TalkingHead 1KH, we only use the portrait images to measure the out-of-domain generative ability of our model.
}
 \vspace{-0.0mm}
	\label{tab:data_mod}
 \scalebox{0.93}{
	\begin{tabular}{l|c|c|c|c}
		\toprule[1.5pt]
        \textbf{Dataset} & \textbf{Task} & \textbf{Video} & \textbf{Audio (Speech)} & \textbf{Instruction} \\
        \midrule
		MEAD & \multirow{2}{*}{Emotinoal Talking Control} & \checkmark & \checkmark &\checkmark \\
		 HDTF &  & \checkmark & \checkmark & \checkmark  (Always neutral) \\
		\midrule
		CC v1         & Facial Motion Control & \checkmark &  & \checkmark\\
  \midrule
		TalkingHead 1KH   & Evaluation* & \checkmark & \checkmark &  \\
		\bottomrule[1.5pt]
	\end{tabular}
 }
\vspace{-0mm}
\end{table}  
}
\subsection{Data Preprocessing}

We preprocess all datasets and establish filter policies to discard low-quality samples.

For video data, we standardize each clip into a portrait-centered talking head video with dimensions of 256 by 256 pixels and a frame rate of 25 fps. Our filtering strategy follows~\cite{he2023gaia} summarized as follows: \textbf{(1)} We maintain consistency in the orientation of individuals facing the camera throughout video clips. Frames exhibiting significant deviations, potentially obscuring lip movements, are excluded. \textbf{(2)} We monitor the positions of faces across frames, ensuring minimal displacement over consecutive timestamps to achieve smooth facial motion in video clips. \textbf{(3)} Frames featuring individuals wearing masks or remaining silent are identified and removed. Additionally, to minimize domain gaps across datasets, we estimate the distribution of talking head position and scale in the HDTF~\cite{zhang2021flow} videos and adjust the other datasets accordingly to this standard.

For each video clip obtained, we extract the audio and resample it to a 16kHz sampling rate. We normalize the speech and apply a denoiser~\cite{defossez2020real} to reduce background noise. In the CC v1 dataset~\cite{hazirbas2021towards}, where the audio comprises off-screen instructional speech unrelated to lip movement, we generate pseudo-empty audio with zero amplitude and a duration matching that of the corresponding ground truth video clip. Subsequently, we extract audio features using Wave2Vec 2.0~\cite{baevski2020wav2vec}.

An important aspect of our model is its integration of textual information as a supervised signal. For emotional talking control, we outline the process of constructing textual instructions in~\cref{sec: method_data} in the main text, along with listing the templates and prompts used in~\cref{sec:app_template}. For facial motion control, the CC v1 dataset provides annotations consisting of off-screen instructional speeches obtained via ASR(Automatic Speech Recognition), along with corresponding timestamps. We extract the instructional annotations and prompt GPT-4V~\cite{gpt4v} to paraphrase them into fluent sentences, eliminating incomplete forms due to ASR detection. Based on these timestamps, we extract the corresponding action videos.
Regarding the HDTF dataset, which lacks explicit instructions, and considering that the majority of videos in this dataset exhibit neutral emotions, we provide pseudo instructions such as ``Talk with neutral emotion'' or ``Talk with an emotionless face''. \etc.

\subsection{Data Statistic}
We provide the statistics of each dataset in~\cref{tab:data_stat}. 
\begin{table}
\centering 
\caption{Statistics of the collected dataset.}
 \vspace{-0.0mm}
	\label{tab:data_stat}
	\begin{tabular}{lcccc}
		\toprule[1.5pt]
		 \multirow{2}{*}{Datasets} & \multicolumn{2}{c}{Raw} & \multicolumn{2}{c}{Filtered} \\
		 \cmidrule(l){2-3}
		 \cmidrule(l){4-5}
		 & \#IDs & \#Hours & \#IDs & \#Hours \\
		\midrule
		HDTF         & $362$ & $16$ & $359$ & $14$  \\
		CC v1 (Motion Control)   & $2,412$ & $29$ & $2,412$ & $23$ \\
            MEAD             & $47$ & $42$ & $47$ & $31$  \\
            \midrule
            Total             & $2,821$ & $87$ & $2,818$ & $68$ \\
		\bottomrule[1.5pt]
	\end{tabular}
\vspace{-0mm}
\end{table}

\subsection{Instruction Templates and GPT-4V Prompt} \label{sec:app_template}

We present a portion of the templates utilized for transforming emotion types into sentences in \cref{tab:inst_temp}, along with the prompts used to query GPT-4V in \cref{tab:gpt4_prompt}.

\section{Implementation Details}\label{sec:app_impl}
\subsection{Illustration of VAE}\label{sec:app_vae}
In our model, we employ a Variational Autoencoder (VAE) based on the framework outlined in~\cite{he2023gaia}. The VAE is designed to disentangle motion information from video data and consists of two encoders: the motion encoder and the appearance encoder, along with a single decoder.

To prevent the leakage of appearance information in reconstruction, they utilize the appearance information from the $i$-th frame and the motion information from the $j$-th frame to reconstruct the $j$-th frame by the VAE. Therefore, in cases where the $i$-th and $j$-th frames from one video clip contain the same appearance but different motion information (e.g., the same person speaking different words), the VAE model learns to first extract the pure appearance feature from the $i$-th frame. Subsequently, it combines this feature with the pure motion feature of the $j$-th frame to accurately reconstruct the original $j$-th frame. 

\update{Once the VAE is well trained, it can encode a video into disentangled appearance and motion latents. For a video with $l$ frames, we can randomly select a frame for appearance encoding, resulting in an appearance latent of shape $(1, d_{app})$, where $d_{app}$ is 768 for the GAIA$\rm{_{base}}$ model. This corresponds to the flattened dimensions $(3, 16, 16)$, capturing spatial information. For the motion latent, we obtain a latent of shape $(l, d_{mot})$ by encoding all frames with the motion encoder, where $d_{mot}$ is also 768.
}

\subsection{Hyperparameters and Model Architectures}

The VAE comprises cascaded traditional convolutional residual blocks for both appearance and motion encoders, with downsampling factors of 8 and 16, respectively. The hidden size and number of layers are set to 256 and 4, resulting in approximately 700M parameters. The motion generator, based on diffusion models~\cite{ho2020denoising}, consists of 12 Conformer~\cite{gulati2020conformer} blocks with a hidden state size of 768. The total number of parameters for the motion generator is 409M, including around 100 million for the non-trainable CLIP-L/14~\cite{CLIP} text encoder and approximately 300 million for the denoising backbone. The adapter~\cite{gao2024clip} utilizes a two-layer MLP with skip connections, where the hidden state size of the middle layer is 4 units smaller than the input dimension.

During training, the Adam~\cite{kingma2014adam} optimizer is employed with $\beta_1=0.9$ and $\beta_2=0.98$. The learning rate starts at 1e-5 and follows an inverse square root schedule with 8000 warmup steps. For the diffusion model, a quadratic $\beta$ schedule is set with $\beta_{min}=0.05$ and $\beta_{max}=20$. During inference, the model follows the DDIM~\cite{song2020denoising} approach and samples 150 steps. The loss weights are set to $\lambda_{pose}=1$, $\lambda_{au}=\lambda_{inten}=0.1$.

\section{Ethical Consideration}\label{sec:app_limit}

For ethical considerations, \modelname is designed to advance AI research on talking avatar generation. Responsible usage is strongly encouraged, and we discourage users from employing our model to generate intentionally deceptive content or engage in other inauthentic activities. To prevent misuse, adding watermarks is a common approach. Moreover, as a generative model, our results can be utilized to construct artificial datasets and train discriminative models.

\begin{table*}[ht!]

\begin{tcolorbox}

{\normalsize \textbf{=========  ~\textsc{Instruction Templates}~ =========}}

\tcblower

"Talk with [EMO] emotion",

"Please talk with a [EMO] expression.", 

"Please talk with a [EMO] face.", 

"Please talk with a [EMO] look.", 

"Please engage in a conversation using an [EMO] expression.",

"Converse with the help of [EMO] expression.",

"Make use of [EMO] expression as we discuss.",

"Could you express [EMO]ly in your speech?",

"speak with the expression of [EMO]",

"Please convey your thoughts with the emotion of [EMO].",

"I'd like you to talk as if you were feeling [EMO].",

"Incorporate [EMO] into your speech.",

"Let your words reflect the sentiment of [EMO].",

"Let [EMO] be your guide in speaking.",

"Express yourself as if you were experiencing [EMO].",

"Speak as if your emotions were [EMO].",

"Try conveying your message with [EMO] as your tone.",

"Let [EMO] be your emotion as you speak.",

"Speak with the emotion of [EMO] guiding your words.",

"Let your words with the feeling of [EMO].",

"I'd like you to communicate using [EMO] expression, please.",

"Could you chat employing [EMO] expression?",

"Discuss the topic while embracing the spirit of [EMO].",

"Let's discuss things while incorporating [EMO] expression.",

"Feel free to use [EMO] expression while we talk.",

"Speak as if you were surrounded by [EMO].",

"Let the essence of [EMO] guide your words.",

"Let [EMO] set the tone for your dialogue.",

"Could you communicate with the spirit of [EMO]?",

"Express your thoughts with the [EMO] in mind.",

"Let [EMO] be the guiding principle in your conversation.",

"Try conveying your message while being [EMO].",

"Let [EMO] set the mood for your conversation.",

"Speak as if your emotions were [EMO].",

"Let [EMO] shape the texture of your conversation.",

"Try conveying your message while being [EMO].",

"Let [EMO] guides your dialogue.",

"Speak as if [EMO].",

"Express your ideas with the emotional of [EMO].",

......

\end{tcolorbox}
\caption{The templates used in transforming emotion types into a sentence.}
\label{tab:inst_temp}
\end{table*}

\begin{table*}[ht!]

\begin{tcolorbox}

{\normalsize \textbf{==========  ~\textsc{Prompt Templates}~ ==========}}

\tcblower

Action unit is a term used in facial expression analysis to describe specific movements of the facial muscles, which can be used to interpret and understand human emotions and expressions, For example, Inner Brow Raiser is the raising of the inner portion of the eyebrows, which is associated with surprise.
~\\

Now I have obtained action units from a video, the form is like this: [``brow\_lowerer'', ``lips\_part'', ``cheek\_raiser''], I want to combine them into a sentence, like ``make brow lower and separated lips, what's more, you can also lift your cheek''.
~\\

More examples like this: [``upper\_lip\_raiser'', ``lips\_part''] -> `` make sure lip raised and lips parted'';
[``brow\_lowerer'', ``lid\_tightener''] -> "drop brow, at the same time you can tighten your lid'' ;
[``lips\_part'', ``lip\_corner\_puller''] -> ``try to part lips, meanwhile stretch lip corner''.
~\\

Now please observe the image \textbf{\textit{<img>}} above, I give predicted action units \textbf{\textit{<au\_list>}}, please turn it into a natural and diverse sentence. If you find a contraction with the image, you can edit the action unit. Give me three examples.
~\\

Pay attention that the AU subject (like brow) should be maintained except for errors while the AU verb (like lower) can be changed. The way you express the sentence can also be free. You should try to make it diverse, clear, and natural but do not imagine an unrealistic subject. 
Avoid using ``you'' if possible. 
Do not use adverbs describing degree, such as ``slightly''. 
Do not incorporate temporal information, such as ``begin'', or ``then''.
~\\

Your answer:

\end{tcolorbox}
\caption{The prompts used to query GPT-4V.}
\label{tab:gpt4_prompt}
\end{table*}
\end{document}